%% file: main.tex
\documentclass[table]{gtech}
\PassOptionsToPackage{table, usenames, dvipsnames}{xcolor}
\usepackage{xcolor}
\usepackage{amsthm}                %
\usepackage{algorithm}      %
\usepackage{algpseudocode}  %
\usepackage{tabularx}
\newcolumntype{C}{>{\centering\arraybackslash}X}

\newtheorem{theorem}{Theorem}[section]
\usepackage{amssymb}
\usepackage{multirow}
\usepackage{bigdelim}
\usepackage{longtable}
\usepackage{tabularray}
\usepackage{wrapfig}
\usepackage[most]{tcolorbox}
\usepackage{url}
\usepackage{float}
\usepackage{datatool}
\usepackage{enumitem}
\usepackage{subcaption} %
\usepackage[justification=centering]{caption}

\RequirePackage{tgpagella} %
\RequirePackage{mathpazo}  %
\RequirePackage{inconsolata} %
\usepackage{makecell}

\renewcommand{\title}[1]{\newcommand{\titlelist}{{\huge\fontfamily{optimistic}\selectfont #1}}}

\definecolor{prompt}{HTML}{5f84e4}
\definecolor{img}{HTML}{820100}

\newcommand{\ignore}[1]{}

\usepackage{arydshln}
\definecolor{CQColor}{rgb}{0.0,0.0,1.0} %

\usepackage{colortbl}
\usepackage{amssymb}
\usepackage{pifont}
\usepackage{booktabs,multirow}
\usepackage{makecell}
\usepackage{tabulary}
\usepackage{fontawesome5}
\usepackage{bbding}
\usepackage{multicol}

\newlength\savewidth

\title{WSM: Decay-Free Learning Rate Schedule \\ via Checkpoint Merging for LLM Pre-training}

\author[1,*]{Changxin Tian}
\author[1,2,*,\dag]{Jiapeng Wang}
\author[1]{Qian Zhao}
\author[1]{Kunlong Chen}
\author[1]{Jia Liu}
\author[1]{Ziqi Liu}
\author[2]{\\Jiaxin Mao}
\author[2,\ddag]{Wayne Xin Zhao}
\author[1,\ddag]{Zhiqiang Zhang}
\author[1]{Jun Zhou}

\affiliation[1]{Ling Team, Ant Group \\ $^2$Gaoling School of Artificial Intelligence, Renmin University of China \\[0.5em]}

\contribution[*]{Equal contribution (listed in alphabetical order)}
\contribution[\dag]{Contribution during internship at Ant Group}

\contribution[\ddag]{Corresponding authors (listed in alphabetical order).}

\date{July 20, 2025\vspace{-1mm}}
\correspondence{\parbox[t]{12cm}{ %
  \raggedright %
  \email{\{tianchangxin.tcx,lingyao.zzq\}@antgroup.com} \\ \email{wangjp1010@ruc.edu.cn}, \email{batmanfly@gmail.com}
}}

\abstract{\fontsize{11pt}{12pt} {Recent advances in learning rate~(LR) scheduling have demonstrated the effectiveness of decay-free approaches that eliminate the traditional decay phase while maintaining competitive performance. Model merging techniques have emerged as particularly promising solutions in this domain. We present Warmup-Stable and Merge (WSM), a general framework that establishes a formal connection between learning rate decay and model merging. WSM provides a unified theoretical foundation for emulating various decay strategies—including cosine decay, linear decay and inverse square root decay—as principled model averaging schemes, while remaining fully compatible with diverse optimization methods. Through extensive experiments, we identify merge duration—the training window for checkpoint aggregation—as the most critical factor influencing model performance, surpassing the importance of both checkpoint interval and merge quantity. Our framework consistently outperforms the widely-adopted Warmup-Stable-Decay (WSD) approach across multiple benchmarks, achieving significant improvements of +3.5\% on MATH, +2.9\% on HumanEval, and +5.5\% on MMLU-Pro. The performance advantages extend to supervised fine-tuning scenarios, highlighting WSM's potential for long-term model refinement.
}}
\addsecondlogo[3.8cm]{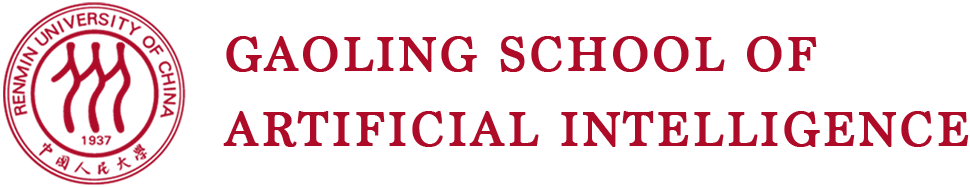} 
\begin{document}
\maketitle

\input{sections/intro}

\input{sections/pre}

\input{sections/method}
\input{sections/exp}

\input{sections/rela}

\input{sections/conclu}

\bibliographystyle{assets/plainnat}
\bibliography{main}

\input{sections/appendix}

\end{document}

%% file: sections/intro.tex
\begin{figure*}[b]
    \centering
    \includegraphics[width=1.\textwidth]{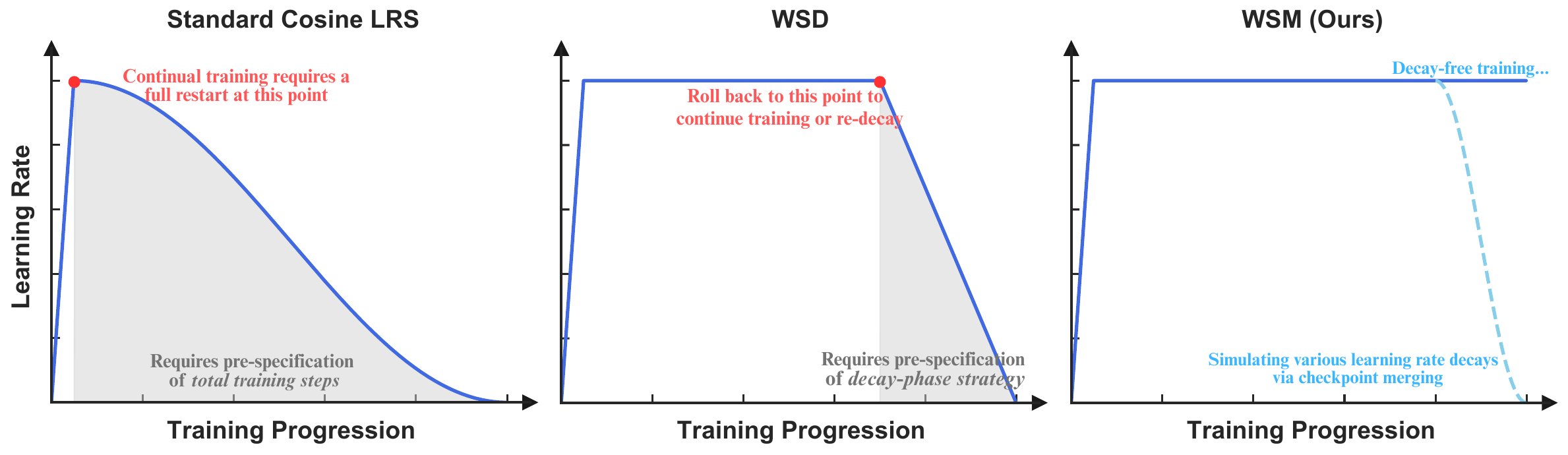}
    \caption{\textbf{Comparison between WSM (Warmup-Stable and Merge) and mainstream learning rate scheduling strategies.}
    By leveraging checkpoint merging, WSM eliminates the learning rate decay phase and maintains a constant learning rate after warmup, enabling a fully autonomous and continuous training process.}
    \label{fig:lrs}
\end{figure*}

\section{Introduction}

In large language model (LLM) pre-training, learning rate (LR) scheduling plays a pivotal role, critically impacting training stability, convergence speed, and final model performance~\citep{Rethinking,Akhilesh}. Conventional LR schedules dynamically adjust the LR based on training progress, with the cosine decay schedule emerging as a widely-adopted approach~\citep{kaplan2020scaling,hoffmann2022training}. This method features an initial warm-up phase followed by cosine-based decay, both constrained by a \emph{predetermined} total training duration. Consequently, any extension of the training process, such as incorporating new data, necessitates a complete restart from the beginning to recalibrate the entire decay curve. 

To address this inflexibility, the Warmup-Stable-Decay (WSD) strategy~\citep{minicpm} inserts a \emph{stable training}  phase with a constant LR between the warmup and decay phases. This schedule provides two key advantages: (1) flexible initiation of the decay phase independent of total step count, and (2) complete decoupling from fixed training durations. The WSD approach has demonstrated significant effectiveness, evidenced by its adoption in several recent LLMs including DeepSeek-V3~\citep{deepseekv3} and  ERNIE 4.5~\citep{ernie2025technicalreport}. %
However, WSD introduces new scheduling requirements: researchers must manually decide when to initiate the decay, how many tokens to allocate for it, and which decay function (e.g., cosine, linear) to employ. Furthermore, if training needs to be extended after the decay has commenced, one must roll back the training to the state preceding the decay phase and re-design the decay strategy. This dependency on a manually configured schedule counteracts the goal of a fully autonomous and continuous training process.

To minimize scheduling complexity, recent research has investigated alternative approaches that completely eliminate the decay phase from LR schedules~\citep{Aaron2024Road,song2025through,Hanlin2025cbs}. A prominent direction in this field explores weight averaging (a.k.a., model merging) techniques. Empirical results demonstrate that simply maintaining a constant LR combined with standard weight averaging strategies—such as exponentially weighted averaging (EWA)—can achieve performance competitive with WSD-based schedules~\citep{pma}. Building upon this line of research, we make three key extensions: 

(1) We present \emph{Warmup-Stable and Merge (WSM)}, a simple yet general framework for LR scheduling. By formalizing the connection between LR decay and checkpoint merging, we demonstrate that WSM can be instantiated to emulate various decay strategies—including cosine decay, linear decay and inverse square root (1-sqrt) decay. Our framework provides a principled approach to convert any LR decay method into a theoretically approximate model averaging implementation. This contrasts with prior work~\citep{Aaron2024Road,song2025through,pma}, which has largely focused on analysis of specific averaging strategies and their optimization properties. Notably, our framework is optimizer-agnostic, enabling seamless integration with various optimization algorithms (e.g., SGD, Adam) without requiring modifications on the underlying training pipeline.

(2) Through extensive experiments, we systematically investigate key factors in instantiating this framework, including merge methods, merge frequency, duration, granularity, and the compatibility of merging and decay strategies. Our findings reveal that merge duration—the training period covered by the merged checkpoints—emerges as the most critical factor influencing model performance, with significantly greater impact than both the checkpoint interval and the number of merged models.

(3) The proposed decay-free LR schedule delivers substantial performance gains. Extensive empirical evaluation shows consistent improvements across multiple benchmarks: +3.5\% on MATH, +2.9\% on HumanEval, and +5.5\% on MMLU-Pro—a significant advance over the WSD method, while prior work often only matches WSD's performance. Moreover, these benefits naturally extend to the post-training stage, demonstrating our method's potential for sustained improvements in long-term model refinement.

The proposed WSM framework presents a promising direction for developing effective decay-free LR schedules. Our results demonstrate consistent performance advantages across both pre-training and fine-tuning stages. Notably, WSM enables the implementation of sophisticated decay-like methods through a simple and stable optimization approach, combining the benefits of strategic scheduling with robust training dynamics.

%% file: sections/pre.tex
\section{Preliminary}
\label{sec:pre}

Typically, existing mainstream LR schedulers consist of two phases: an initial warm-up followed by decay. 
During the warm-up, the LR increases linearly from a small value to a peak value, $lr_{peak}$, over $T_{warmup}$ steps. This helps stabilize the optimization process in the early stages of training. Following the warm-up, the LR gradually decreases according to a predefined function, such as cosine, linear, or inverse square root decay. These schedules share a critical limitation: they require the total number of training tokens (or steps), $T_{max}$, to be known in advance.
For instance, the prevalent cosine LR schedule is formulated as:
\[
lr(t) = 
\begin{cases}
    lr_{peak} \cdot \frac{t}{T_{warmup}} & \text{if } t < T_{warmup} \\
    \frac{1}{2} lr_{peak} \left(1 + \cos\left(\frac{\pi(t - T_{warmup})}{T_{max} - T_{warmup}}\right)\right) & \text{if } t \ge T_{warmup}
\end{cases}
\]

The Warmup-Stable-Decay (WSD) LR schedule introduces a \emph{stable} phase between warm-up and decay, maintaining a constant LR at its peak (\(lr_{peak}\)), which is defined as:
\[
lr(t) = 
\begin{cases}
    lr_{peak} \cdot \frac{t}{T_{warmup}} & \text{if } t < T_{warmup} \\
    lr_{peak} & \text{if } T_{warmup} \le t < T_{decay\_start} \\
    \operatorname{decay\_function}(t) & \text{if } T_{decay\_start} \le t \le T_{max}
\end{cases}
\]
As shown, WSD eliminates the need for manual LR tuning during the stable phase while supporting multiple decay attempts from its endpoint—without requiring a reset to the initial state. 
Despite this flexibility, it still requires predefined decay-phase settings—such as the decay start step ($T_{decay\_start}$), decay function ($\operatorname{decay\_function}$), and total training steps ($T_{max}$). %

%% file: sections/method.tex
\section{The Proposed Methodology}

In this section, we first establish the theoretical connection between checkpoint merging and LR decay, formalize our proposed WSM (Warmup-Stable and Merge) schedule, and compare it with the widely used WSD schedule.

\subsection{Theoretical Connection Between LR Decay and Checkpoint Merging}
\label{Theoretical}
The core idea of checkpoint merging in this work is to take an ordered list of checkpoints, $[\theta_n, \theta_{n+1}, \dots, \theta_{n+k}]$, and apply a merge function to generate a single model $\hat{\theta}_{n+k}$. Here, $\theta_i \in \mathbb{R}^d$ represents the model's parameter vector at the $i$-th training iteration.
The most general form is a weighted average of the checkpoints:
\begin{equation}
    \hat{\theta}_{n+k} = \sum_{j=0}^{k} c_j \theta_{n+j}
    \label{eq:weighted_avg}
\end{equation}
where $\{c_j\}$ are non-negative weights that sum to one, i.e., $\sum_{j=0}^{k} c_j = 1$.

This formulation obscures a deeper connection to the training dynamics. We can reveal this connection by expressing each checkpoint in terms of an initial checkpoint $\theta_n$ and the subsequent gradient updates. 
For simplicity, we assume the updates between checkpoints at different time steps are independent and ignore optimizer states. Let $g_{i}$ be the gradient update vector (including the LR) at step $i$, such that the model updates as $\theta_{i+1} = \theta_i - g_i$. Thus, an intermediate checkpoint $\theta_{n+j}$ can be expressed as the sum of an initial state $\theta_n$ and the sequence of negative gradient updates that followed:
\begin{equation}
    \theta_{n+j} = \theta_{n} - \sum_{l=1}^{j} g_{n+l-1}
    \label{eq:theta_as_gradient_sum}
\end{equation}
\begin{figure*}[t]
    \centering
    \includegraphics[width=\textwidth]{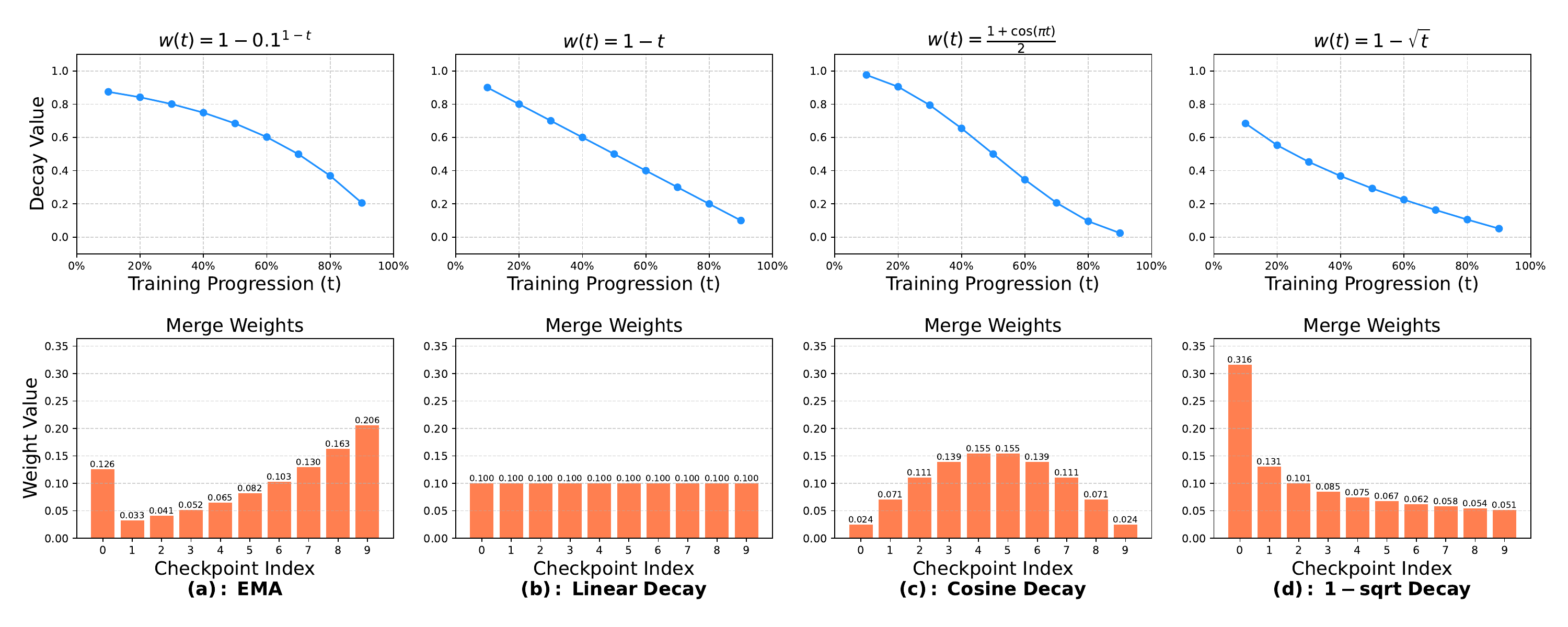}
    \caption{\textbf{
    Visualization of checkpoint merging weight distributions and their corresponding decay functions over a span of 10 checkpoints (larger checkpoint indices denote more recent checkpoints).}
(a) Exponential Moving Average (EMA) weights, exhibiting convex decay characteristics;
(b) Uniform averaging weights, demonstrating linear decay behavior;
(c) and (d) Our theoretically derived weights, designed to approximate cosine and 1-{sqrt} decay patterns.
Note: Decay curves are rendered smoothly for clarity, though the underlying weights are applied discretely at each checkpoint.
}
    \label{fig:weight}
\end{figure*}

Furthermore, we can substitute Eq.~\ref{eq:theta_as_gradient_sum} into the general merging formula (Eq. \ref{eq:weighted_avg}) and then rearrange the double summation by changing its order. 
A gradient update $g_{n+i-1}$ is included in the sum for all involved checkpoints $\theta_{n+j}$ where $j \ge i$. 
\begin{align}
    \hat{\theta}_{n+k} &= \sum_{j=0}^{k} c_j \left( \theta_n - \sum_{l=1}^{j} g_{n+l-1} \right) = \theta_n - \sum_{i=1}^{k} \left( \sum_{j=i}^{k} c_j \right) g_{n+i-1} \label{eq:rearrange_sum_step1}
\end{align}
This shows that a weight coefficient of $\sum_{j=i}^{k} c_j$ is applied to the gradient update $g_{n+i-1}$ from step $n+i-1$.
By defining a new set of weights for the gradient updates, $w_i = \sum_{j=i}^{k} c_j$, we arrive at the final equivalent expression for checkpoint merging:
\begin{equation}
    \hat{\theta}_{n+k} = \theta_n - \sum_{i=1}^{k} w_i \cdot g_{n+i-1}
    \label{eq:merge_as_gradient_reweighting}
\end{equation}
This equation shows that merging checkpoints with weights ${c_j}$ is equivalent to applying a synthetic decay schedule defined by weights ${w_i}$ to the gradients accumulated after the base checkpoint $\theta_n$, where a mapping exists between the merge weights ${c_j}$ and the effective learning rates ${w_i}$.
We therefore propose Theorem~\ref{thm:inverse}, which can approximate monotonically decreasing decay curve. Figure~\ref{fig:weight} illustrates various checkpoint merging weights and their corresponding decay curves. Additional proof details are provided in Appendix~\ref{app:Theoretical}.

\begin{theorem}[Checkpoint Weight Derivation from Gradient Decay Schedule]
\label{thm:inverse}
Given a desired sequence of gradient decay coefficients $\{w_i\}_{i=1}^k$ that is monotonically non-increasing and bounded, $1 \ge w_1 \ge w_2 \ge \dots \ge w_k \ge 0$, the corresponding non-negative checkpoint weights $\{c_j\}_{j=0}^k$ required to satisfy Eq. \ref{eq:merge_as_gradient_reweighting} are uniquely determined by:
\begin{equation}
\begin{cases}
    c_k = w_k \\
    c_j = w_j - w_{j+1}, & \text{for } j \in [1, k-1] \\
    c_0 = 1 - \sum_{j=1}^{k} c_j = 1 - w_1
\end{cases}
\label{eq:inverse_formula}
\end{equation}
\end{theorem}

\subsection{Emulating LR Decay through Checkpoint Merging}
The central hypothesis of WSM is that \textit{the optimization benefits of LR decay can be decoupled from the live training process and instead be effectively achieved through the merging of model checkpoints}~\citep{adam,deepseekv3,pma}.
The WSM simplifies the LR schedule by completely eliminating the decay phase:
\[
lr(t) = \begin{cases}
lr_{peak} \cdot \frac{t}{T_{warmup}} & \text{if } t < T_{warmup} \\
lr_{peak} & \text{if } t \ge T_{warmup}  \\
\end{cases}
\]
The WSM pipeline, detailed in Algorithm~\ref{alg:WSM-anneal}, operates in two primary phases. It begins with a standard warmup phase, where the learning rate linearly increases to its peak value, $lr_{peak}$. Following this, the process enters the main stable training phase, where the learning rate is held constant. After a specified step $T_{switch}$, the model can transition from the general pre-training data $D$ to a smaller, high-quality annealing dataset $D_{anneal}$. This allows the ``annealing'' to be focused on a curated data mixture, aligning with practical pre-training methodologies. Throughout this stable phase, checkpoints are periodically saved. Concurrently, an asynchronous merging process continuously fetches the last $n$ checkpoints from storage and combines them into model $W_{merged}$. Specifically, for the $\operatorname{Merge}(\cdot)$ operation, we can select various decay strategies to emulate (e.g., the decay curve shape and minimum LR), calculate the corresponding gradient decay coefficients $\{w_i\}$, and then derive the checkpoint merging weights $\{c_i\}$ based on Theorem \ref{thm:inverse}. This merged checkpoint, which emulates the effect of a decay schedule, provides a robust, annealed model without ever altering the live learning rate.

\begin{algorithm}[H] %
\small
\caption{The WSM LRS Pre-training Pipeline}
\label{alg:WSM-anneal}
\begin{algorithmic}
\State \textbf{Input:} Initial model weights $W_0$, pre-training data $D$, high-quality annealing data $D_{anneal}$ (optional), peak LR $lr_{peak}$, warmup steps $T_{warmup}$, switch step $T_{switch}$, checkpointing interval $T_{cpt}$, merging window size $n$
\Statex %
\State \textbf{Initialize:}
\State $\triangleright$ The merged checkpoint.
\State $W_{merged} \gets \text{null}$ 
\State $\triangleright$ Storage for the base (un-merged) checkpoints.
\State $\mathcal{C}_{storage} \gets []$ 
\Statex
\State $\triangleright$ \textbf{Phase 1: Warmup on Pre-training Data}
\For{$t = 1$ to $T_{warmup}$}
    \State $lr(t) \gets lr_{peak} \cdot (t / T_{warmup})$
    \State $W_t \gets \text{Update}(W_{t-1}, D, lr(t))$
\EndFor
\Statex
\State $\triangleright$ \textbf{Phase 2: Stable Training and Merging}
\For{$t = T_{warmup} + 1$ to ...}
    \State $lr(t) \gets lr_{peak}$
    \State $\triangleright$ Select dataset for the current step
    \If{$t > T_{switch}$ \textbf{and} $D_{anneal}$ is available}
        \State $D_{current} \gets D_{anneal}$
    \Else
        \State $D_{current} \gets D$
    \EndIf
    \State $W_t \gets \text{Update}(W_{t-1}, D_{current}, lr(t))$
    \If{$t \pmod{T_{cpt}} == 0$}
        \State $\triangleright$ Save the checkpoint from the main training process.
        \State $\text{SaveToCheckpointStorage}(\mathcal{C}_{storage}, W_t)$
        
        \State $\triangleright$ \textbf{Asynchronously Update the Merged checkpoint}
        \State $C_{latest} \gets \text{GetLastNCheckpoints}(\mathcal{C}_{storage}, n)$
        \If{$\text{len}(C_{latest}) == n$}
            \State  $\triangleright$ Update the merged checkpoint for evaluation.
            \State $W_{merged} \gets \text{Merge}(C_{latest})$ 
        \EndIf
    \EndIf
\EndFor

\State \textbf{return} The stored base checkpoints in $\mathcal{C}_{storage}$ and the merged checkpoint $W_{merged}$.
\end{algorithmic}
\end{algorithm}
\subsection{Discussion}

\paragraph{Choice between offline and online merging}
Initially, we employ an offline, checkpoint-based approach as a powerful exploration framework to discover optimal annealing strategies without impacting the primary training run. By preserving a history of all discrete checkpoints, practitioners can systematically evaluate how different factors influence the final model, including the choice of merging algorithm (simulating various decay curves like linear or cosine) and the annealing duration (by varying the merge window size, $n$). This multi-faceted exploration is impossible with a standard online method, such as an Exponential Moving Average (EMA), which hard-codes a single annealing path into the training process. Crucially, once this offline exploration identifies a superior, fixed strategy (e.g., ``an average of the last 4 checkpoints''), it can then be operationalized as a simple and efficient online process using a sliding window.

\paragraph{Practicality and complexity}
Compared to discovering an optimal decay schedule through multiple, resource-intensive training runs, the merging operations in the WSM strategy save substantial computational resources. The main complexity introduced is storage overhead. For the offline merging approach, which is ideal for exploration, one must store a history of checkpoints, with the total number being the total training tokens divided by the checkpointing interval. However, this is a manageable trade-off, as this storage footprint represents a minor fraction of a typical pre-training budget. Furthermore, once a superior strategy is identified, or in scenarios with extreme storage constraints, an online merging approach can be used. This method minimizes the storage footprint by maintaining only a small, fixed-size window of $n$ checkpoints. Our experiments confirm that a relatively small window (e.g., $n=12$) is sufficient to achieve strong results.

\paragraph{Comparison with \cite{pma}.} 
We notice a concurrent work~\citep{pma} studied model merging in pre-training. We highlight the differences between our work and theirs as follows: 
(1) \textbf{Motivation}: We frame checkpoint merging as a novel LR scheduling mechanism, with the primary goal of discovering a schedule that outperforms strong baselines like WSD. In contrast, their work treats model merging as a standalone pre-training technique, evaluating its value by comparing the performance of the merged model against its before-merge checkpoints.
(2) \textbf{Methodology}: Our approach is theory-driven. We first establish a formal theoretical connection between LR decay and checkpoint merging (Section~\ref{Theoretical}) and then leverage merge weights to simulate various decay curves. They mainly conducts empirical studies of merging algorithms, proposing heuristic methods including WMA, SMA, and EMA. 
(3) \textbf{Key Findings}: Through extensive experiments, we identify the merge duration as the most critical hyperparameter influencing final model performance. Ultimately, our method \textit{surpasses} the performance of WSD-annealed baseline (Section~\ref{sec:main}). In contrast, their work concludes that checkpoint merging can effectively \textit{match} the performance of an annealed model. 
The findings from our work and that of \cite{pma} are complementary, collectively providing practical insights into checkpoint merging for LLM pre-training.

\section{Experiment}
\label{sec:exp}

\subsection{Experiment Setup}
The model we used for the experiment is Ling-mini, a standard MoE model with a total of 16.3 billion parameters and 1.4 billion active parameters. Ling-mini adopts a fine-grained expert configuration, consisting of 1 dense layer and 19 MoE layers. Each MoE layer includes 256 experts, with 8 experts being activated for each token, along with one additional shared expert. We utilized the AdamW optimizer~\citep{adamw}, and the hyperparameters are set to $\beta_1 = 0.9$ and $\beta_2 = 0.95$, with 0.1 weight decay applied. Through preliminary scaling laws experiments, we set the peak LR and batch size to 4.78e-4 and 2048, respectively.

We begin with a checkpoint pretrained on 10.2 trillion tokens total - comprising 10 trillion tokens at standard 4,096 length sequences plus 200 billion long-context tokens - all trained using a stable, constant LR. 
Then, we introduce a specialized high-quality annealing dataset and branch the training into two distinct strategies for an additional 400B tokens to compare their effectiveness:
(1) We apply a conventional LR decay schedule to the model. This branch serves as our baseline, representing the standard Warmup-Stable-Decay (WSD) methodology.
(2) We continue training with the same constant LR. The final model is then produced by merging the checkpoints saved during this stable phase. This branch represents our proposed WSM schedule.
Unless otherwise specified, we save a checkpoint every 25B tokens and use mean averaging to merge the most recent checkpoints.
Comprehensive details regarding our model architecture, specific training parameters, dataset composition, and evaluation protocols are provided in Appendix~\ref{setup}.

\subsection{Overall Performance of WSM Schedule}
\label{sec:main}
\begin{figure*}[t]
    \centering
    \includegraphics[width=\textwidth]{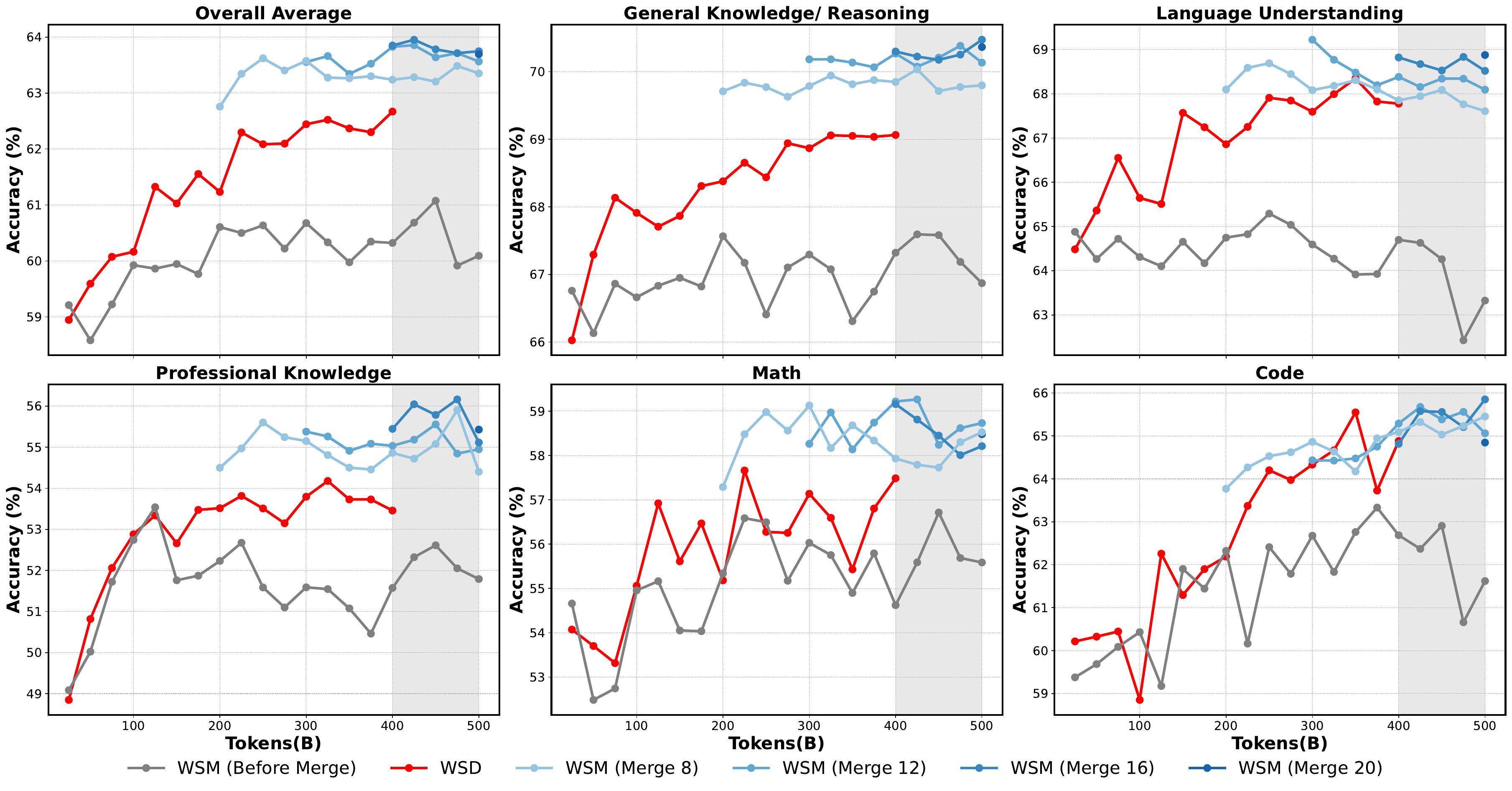}
    \caption{\textbf{Comprehensive performance comparison (overall and by category) between our WSM schedule (via checkpoint merging) and standard WSD scheduling (via learning rate decay).} Both approaches are initialized from the same pretrained checkpoint (10.2T tokens with constant LR). Notably, while WSD requires predetermined decay strategy (e.g., decay over 400B tokens in this study), WSM eliminates such constraints, enabling seamless training continuation (gray regions) and flexible decay behavior approximation.}
    \label{fig:main}
\end{figure*}

\paragraph{Immediate effects during pre-training }
In this section, we present a comprehensive performance comparison between our proposed WSM schedule and the baseline WSD schedule. We evaluate three series of checkpoints: (1) those obtained using the standard LR decay schedule in WSD, (2) checkpoints from the last stable phase of WSM before merging, and (3) our final merged checkpoints from the WSM method using various merge durations (window sizes). The category-wise average results are summarized in Figure~\ref{fig:main} and Table~\ref{tab:base_model_comparison}, with scores for each benchmark provided in Appendix~\ref{all_datasets}. 

Our first and most significant finding is that the WSM method consistently outperforms WSD across the majority of tasks considered. On average, WSM achieves performance improvements across all benchmark categories.
Notably, when comparing the best-performing checkpoint from each strategy, WSM improves upon WSD by an average of 1.3 points.
We observe remarkable improvements of up to 2.7 points on MATH, 2.4 points on HumanEval, 2.1 points on MMLU-Pro.
These results provide compelling evidence that replacing the LR decay phase used in WSD with the checkpoint merging strategy of WSM is not only a feasible alternative but also a more effective approach for enhancing the diverse capabilities of the final pretrained model.

\paragraph{Long-term implications for post-training}
While WSM demonstrates promising results in the pre-training phase, we further investigate its long-term impact on subsequent post-training stages. We apply supervised fine-tuning on checkpoints generated by the WSM and WSD under identical settings for 5 epochs. Results in Table~\ref{tab:instruct_model_comparison} show that this advantage persists beyond the post-training phase.

\begin{table}[htbp]
  \centering
  \small
  
  \caption{Base model performance comparison. Results are reported based on the checkpoint with the highest average benchmark score.}
  \label{tab:base_model_comparison}
  \begin{tabularx}{\textwidth}{ c l C C C C C c C }
    \toprule
    & & General Knowledge & Language Modeling & Math & Code & Professional Knowledge & & Overall Average \\
    \midrule
    \multirow{3}{*}{Base Model}
    & WSD                & 69.06 & 67.78 & 57.49 & 64.88 & 53.46 & & 62.67 \\
    & WSM                & \textbf{70.22} & \textbf{68.67} & \textbf{58.81} & \textbf{65.58} & \textbf{56.04} & & \textbf{63.95} \\ \cmidrule(l){2-9} 
    & Improv.            & +1.68\% & +1.31\% & +2.30\% & +1.08\% & +4.83\% & & +2.04\% \\ 
    \bottomrule
  \end{tabularx}

  \vspace{0.3cm} %

  \caption{Instruct model performance comparison. Results are reported based on the epoch with the highest average benchmark score.}
  \label{tab:instruct_model_comparison}
  \begin{tabularx}{\textwidth}{ c l C C C C C C c C }
    \toprule
    & & Language & Knowledge & Math & Code & Reason & Agent & & Overall Average \\
    \midrule
    \multirow{3}{*}{Inst Model}
    & WSD     & 81.12 & 60.00 & 61.43 & \textbf{58.23} & 63.21 & 68.16 & & 62.90 \\
    & WSM     & \textbf{84.78} & \textbf{61.73} & \textbf{62.28} & 57.95 & \textbf{64.94} & \textbf{69.33} & & \textbf{64.07} \\ \cmidrule(l){2-10} 
    & Improv. & +4.51\% & +2.88\% & +1.38\% & -0.48\% & +2.74\% & +1.72\% & & +1.86\% \\
    \bottomrule
  \end{tabularx}
\end{table}

%% file: sections/exp.tex
\subsection{Empirical Analysis of WSM Schedule}
In this section, we conduct a comprehensive empirical study to dissect the WSM schedule. We aim to understand the key factors influencing its performance, its robustness across different training scenarios, its interaction with conventional decay mechanisms, and its broader implications on model dynamics.

\subsubsection{Robustness Across Pre-training Process}
Beyond applying WSM as a final step on a high-quality dataset, we also evaluated its utility and robustness throughout the entire pre-training lifecycle. To achieve this, we conducted a comparative analysis at various intermediate stages of a long training run, comparing the performance of a model produced by our computationally-frugal WSM merging against one produced by initiating a full, resource-intensive learning rate (LR) decay.
As shown in Figure \ref{fig:weight_a}, although the performance gains over WSD are not as significant as when switching to high-quality data, we find that the performance of models generated via WSM merging consistently and closely mirrors the results of a true LR anneal. In the figure, the gray line represents the base model trained with a constant LR. The blue lines show the performance of WSM models, which were created by mean-merging four checkpoints from within the preceding 100B-token merge duration. The red lines represent full 100B-token decay runs initiated at the 2T, 4T, 6T, 8T, and 10T token milestones. This result establishes WSM as a reliable, high-fidelity proxy for estimating a model's post-anneal potential at any point during training. Consequently, it can provide effective assessment throughout the pre-training phase, eliminating the need to launch multiple, expensive decay runs to gauge the model's true strength.

\subsubsection{Impact of Merging Algorithm}
As derived in Section~\ref{Theoretical}, the checkpoint merging process can be viewed as an approximation of a LR decay schedule, where the weighting scheme of the merge is analogous to the functional form of the decay curve. For instance, a simple mean average is analogous to a linear decay curve. An Exponential Moving Average (EMA) would correspond to a convex exponential decay curve.
Existing works~\citep{1sqrt} and our prior experiments with the WSD schedule revealed a performance hierarchy among decay curves: concave schedules (e.g., inverse square root) and linear schedules outperform convex schedules (details are provided in Appendix~\ref{app:add_exp}). Building on these findings, we hypothesize that a merging algorithm designed to approximate decay curves that have been proven effective in WSD scheduling will similarly yield better results.

Based on Theorem~\ref{thm:inverse}, we experimentally compare three merging algorithms: one using our theorem-generated weights to approximate 1-\textit{sqrt} decay, another using simple mean averaging, and a third using EMA.
Our experimental results in Table~\ref{tab:merge_comparison_fixed} validate this hypothesis. While the merge method outperforms decay, the 1-\textit{sqrt} merge approach shows slight advantages over Mean, and both are markedly better than EMA. This finding reinforces the theoretical connection between checkpoint merging and LR decay, suggesting that the benefits of superior decay schedules can be effectively captured through carefully designed merging weights.

\subsubsection{Impact of Merging Duration and Granularity}
\begin{figure*}[t]
    \centering
    \includegraphics[width=1\textwidth]{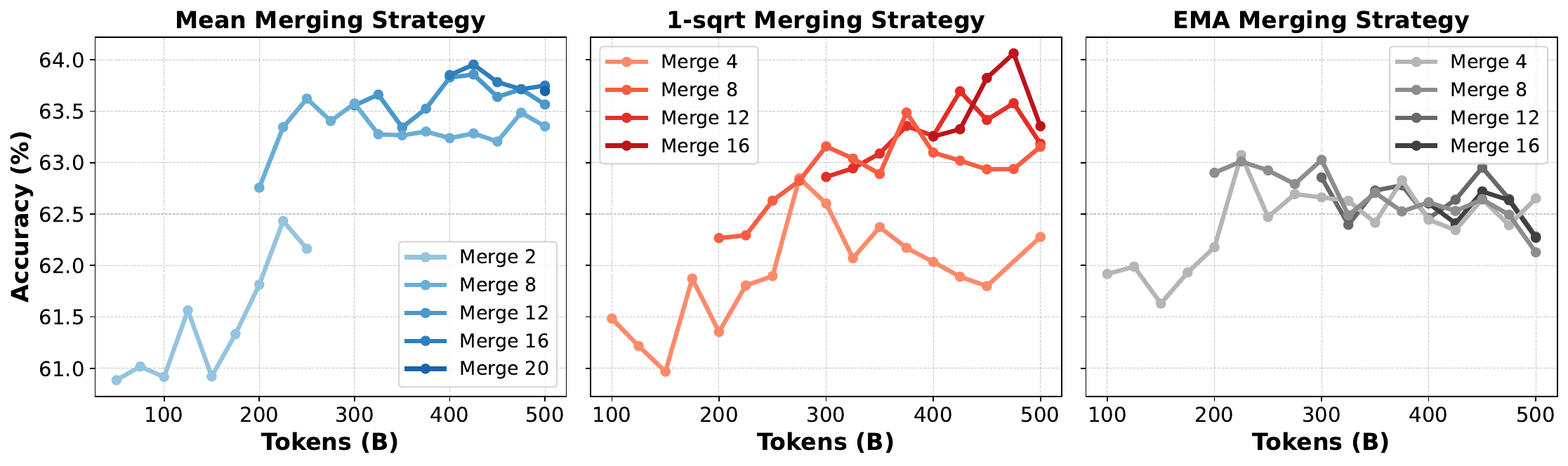}
    \caption{\textbf{Merge duration analysis across algorithms.}``Merge 4'' indicates merging the most recent 4 checkpoints, with each checkpoint saved at 25B-token intervals.}
    \label{fig:window}
\end{figure*}

We further investigate in detail the impact of different merge durations (window sizes) on various algorithms.
First, different merge durations essentially correspond to decaying with varying amounts of data.
As shown in Figure \ref{fig:window}, when comparing the best-performing checkpoints across the entire merging trajectory for both mean and 1-\textit{sqrt} merging algorithms, larger merging windows tend to yield better results.
However, this advantage gradually diminishes as the window size increases. This observation aligns with previous practical LR decay experiments, where simply increasing the amount of annealing data shows diminishing returns and may eventually fail to provide further improvement. These findings further strengthen the connection between LR decay and checkpoint merging.
For EMA merging, its performance is significantly inferior to other algorithms and shows no clear variation with merge durations. This may indicate that EMA is not an effective merging algorithm (also suggesting that the convex nature may not represent an optimal decay curve).

Then we further investigate the granularity of checkpoint merging, i.e., the interval between saved checkpoints used for merging. Finer-grained merging represents a more precise approximation of the true decay curve. The results in Table~\ref{tab:Granularity} show that finer-grained merging tends to achieve better performance. However, frequent checkpoint saving imposes storage overhead, requiring careful trade-off considerations.

\begin{table}[htbp]
  \centering
  \small
  \caption{\textbf{Impact of merging algorithm.} }
  \label{tab:merge_comparison_fixed}
  \begin{tabularx}{\textwidth}{ c c C C C C C C }
    \toprule
    &  & General Knowledge & Language Modeling & Math & Code & Professional Knowledge & Overall Average\\
    \midrule
    \multirow{1}{*}{Decay}
    &  1-\textit{sqrt}     & 69.06 & 67.78 & 57.49 & 64.88& 53.46 & 62.67 \\
    \midrule
    \multirow{3}{*}{Merging}
    & EMA                 & 69.05 & 67.64 & 58.81 & 64.19 & 54.44 & 63.01 \\
    & Mean                & 70.22 & \textbf{68.67} & 58.81 & 65.58& \textbf{56.04} & 63.95 \\
    &  1-\textit{sqrt}       & \textbf{70.27} & 68.26 & \textbf{59.65} & \textbf{65.70} & 55.42 & \textbf{64.06} \\
    \bottomrule
  \end{tabularx}
\end{table}

\begin{table}[htbp]
  \centering
  \small
  \caption{\textbf{Comparison of different saving/merging intervals within an 80B-token merge duration.} For example, (5B,16) indicates that saving every 5B tokens while merging the latest 16 checkpoints.}
  \label{tab:Granularity}
  \begin{tabularx}{\textwidth}{ C c C C C C C C }
    \toprule
    Merging Granularity & & General Knowledge & Language Modeling & Math & Code & Professional Knowledge & Overall Average\\
    \midrule
    (5B,16)                 & & \textbf{69.46} & 79.95 & 57.07 & \textbf{64.68} & 53.82 & 63.63 \\
    (10B,8)                & & 69.23 & 80.00 & \textbf{57.94} & 64.39& 53.78 & \textbf{63.78} \\
    (20B,4)                & & 69.29 & \textbf{80.29} & 56.79 & 63.87 & \textbf{54.19} & 63.36 \\
    (40B,2)                & & 68.47 & 79.83 & 56.57 & 63.59 & 52.20 & 62.77 \\
    (80B,1)                & & 67.47 & 64.98 & 55.07 & 61.69 & 51.61 & 60.33 \\
    \bottomrule
  \end{tabularx}
\end{table}

\subsubsection{On the Compatibility of Merge and Decay}
Given that checkpoint merging effectively simulates LR decay, a natural question arises: can merging and decay be combined to achieve synergistic performance gains? We investigate this by testing two hybrid approaches. (1) Decay-then-Merge: We first apply a standard decay schedule and then merge checkpoints selected from within the decay phase. (2) Merge-then-Decay: We further apply a decay schedule to the resulting merged model.
\begin{figure*}[t]
    \centering
    \begin{subfigure}[t]{0.32\textwidth}
        \includegraphics[width=\linewidth]{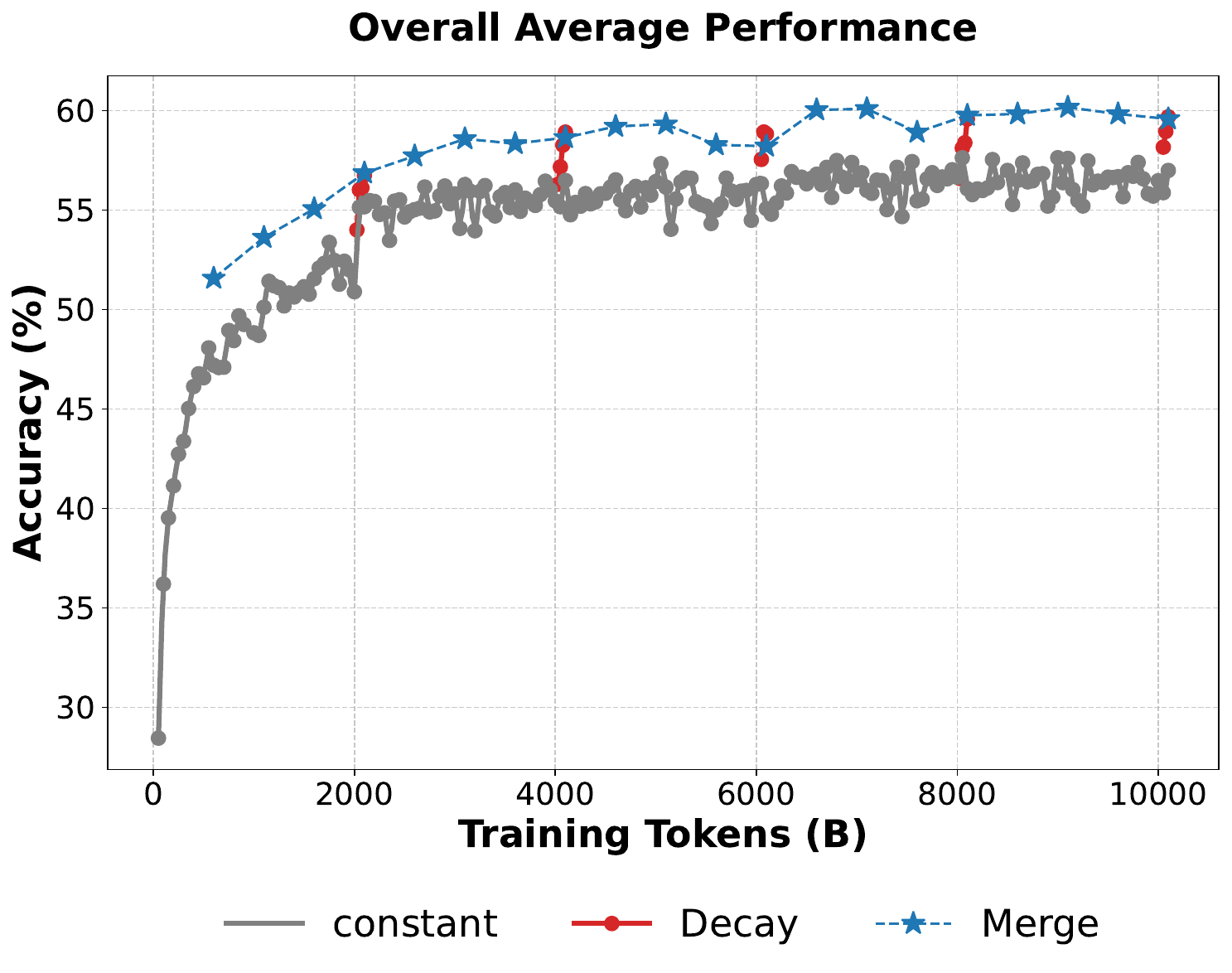}
        \caption{}
        \label{fig:weight_a}
    \end{subfigure}
    \hfill
    \begin{subfigure}[t]{0.32\textwidth}
        \includegraphics[width=\linewidth]{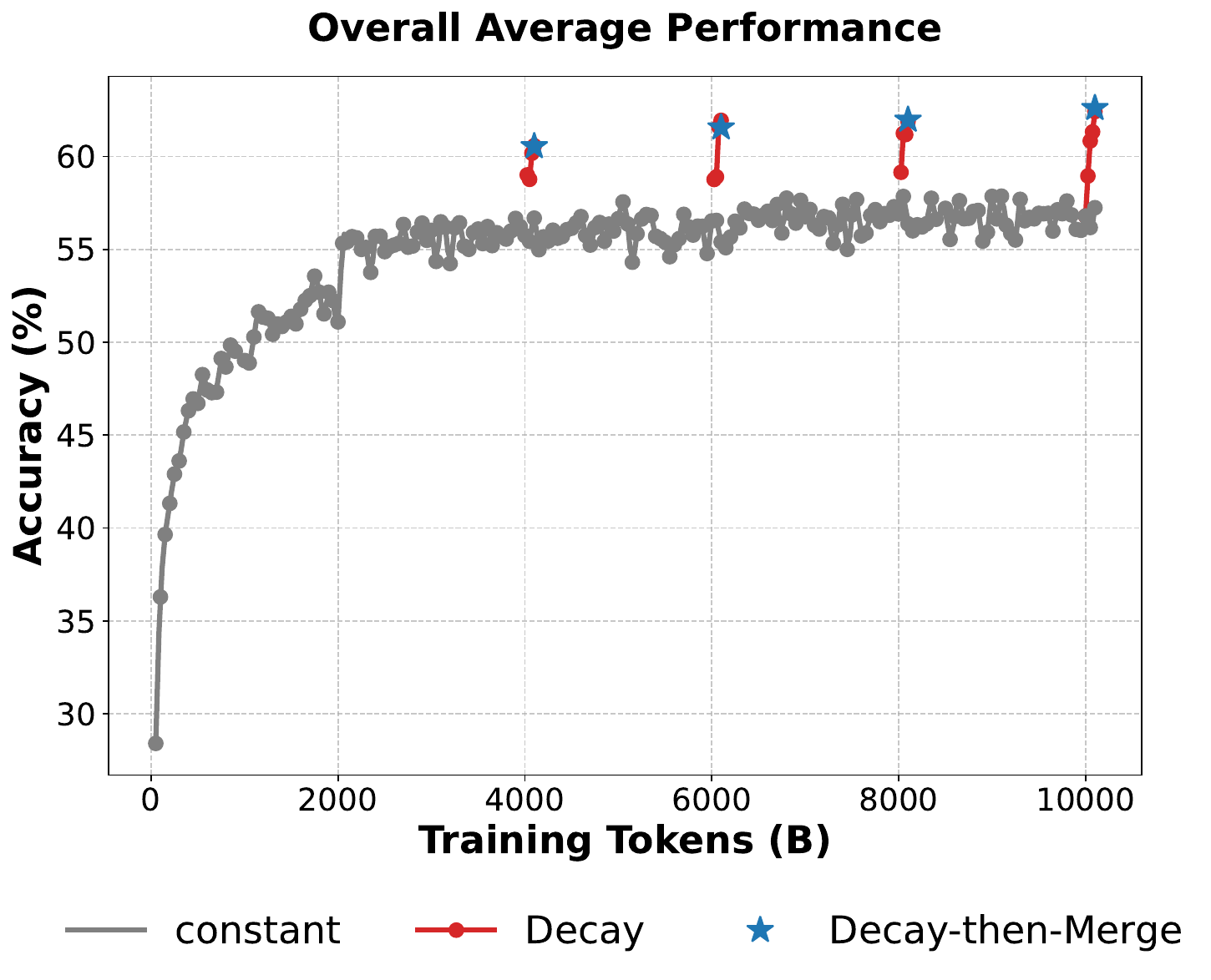}
        \caption{}
        \label{fig:weight_b}
    \end{subfigure}
    \hfill
    \begin{subfigure}[t]{0.32\textwidth}
        \includegraphics[width=\linewidth]{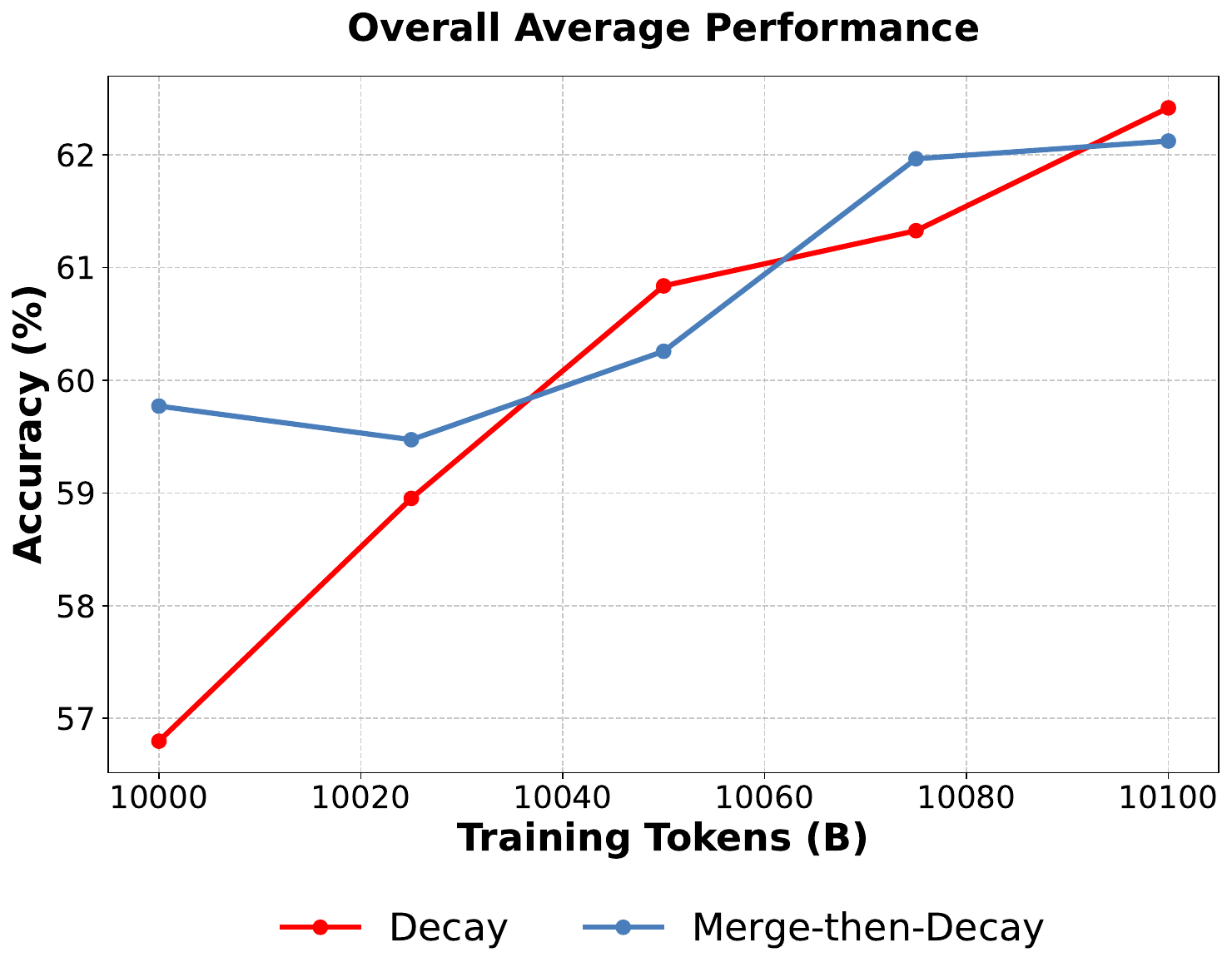}
        \caption{}
        \label{fig:weight_c}
    \end{subfigure}
    \caption{(a) A comparison of checkpoint merging (WSM) and true LR decay (WSD) across a long pre-training run. (b) Investigating the effect of applying merging within a decay phase (Decay-then-Merge). (c) Investigating the effect of applying LR decay after checkpoint merging (Merge-then-Decay).}
    \label{fig:all}
\end{figure*}
As shown in Figures \ref{fig:weight_b} and \ref{fig:weight_c}, the hybrid approach failed to yield any improvement in either configuration, although the Merge-then-Decay model showed better performance at the beginning of its training. For the Decay-then-Merge experiment (Figure \ref{fig:weight_b}), the blue stars represent the results of merging checkpoints selected along the decay trajectories (red lines), which were initiated at various pre-training milestones (4T, 6T, 8T, 10T). For the Merge-then-Decay experiment (Figure \ref{fig:weight_c}), we compare a decay run initiated from a WSM model (blue line)—created by merging four checkpoints from the 9.8T to 10T token interval—against a standard decay baseline initiated from a single 10T checkpoint (red line). These results suggest that checkpoint merging and LR decay are not complementary but rather alternative pathways to a similar optimization objective.

\subsubsection{Impact on MoE Load Balancing}
\begin{table}[htbp]

\centering
\caption{\textbf{Impact on MoE load balancing.} The WSM strategy demonstrates improved expert utilization (lower load balancing violation scores) with a slightly higher language modeling loss.}
\begin{tabular}{lccc}
\toprule
 & language modeling loss & mean\_global\_max\_violation & mean\_global\_min\_violation \\
\midrule
WSD & 0.675 & 0.601 & 0.322 \\
WSM & 0.697 & \textbf{0.545} & \textbf{0.201} \\
\bottomrule
\end{tabular}
\label{route}
\end{table}

We provide a more in-depth analysis of the implications of WSM on MoE router balance in Table~\ref{route}. Specifically, the violation for a single expert is calculated as its relative deviation from the average load within its layer\footnote{For a single layer, let $L$ be the vector of token loads for its experts and $\mu = \text{mean}(L)$, $\text{max\_violation} = \frac{\lvert \max(L) - \mu \rvert}{\mu}$ and $\text{min\_violation} = \frac{\lvert \min(L) - \mu \rvert}{\mu}$}. The mean\_global\_max\_violation represents the average of the highest violations across all layers (measuring the severity of ``overloaded'' experts), while mean\_global\_min\_violation averages the violations for the least-utilized experts (measuring the risk of ``routing collapse'').
When comparing the merged checkpoint with a decayed checkpoint, the merged checkpoint achieves more balanced routing, although its loss is slightly higher.
We argue that this trade-off—a marginally higher loss for superior downstream performance—is indicative of enhanced generalization rather than overfitting to the training data.
More analysis of the parameter distribution is provided in Appendix \ref{parameter}.

\subsection{Parameter Trajectories and Model Performance: Constant LR with Merge vs. LR Decay}
\begin{figure*}[t]
    \centering
    \includegraphics[width=0.8\textwidth]{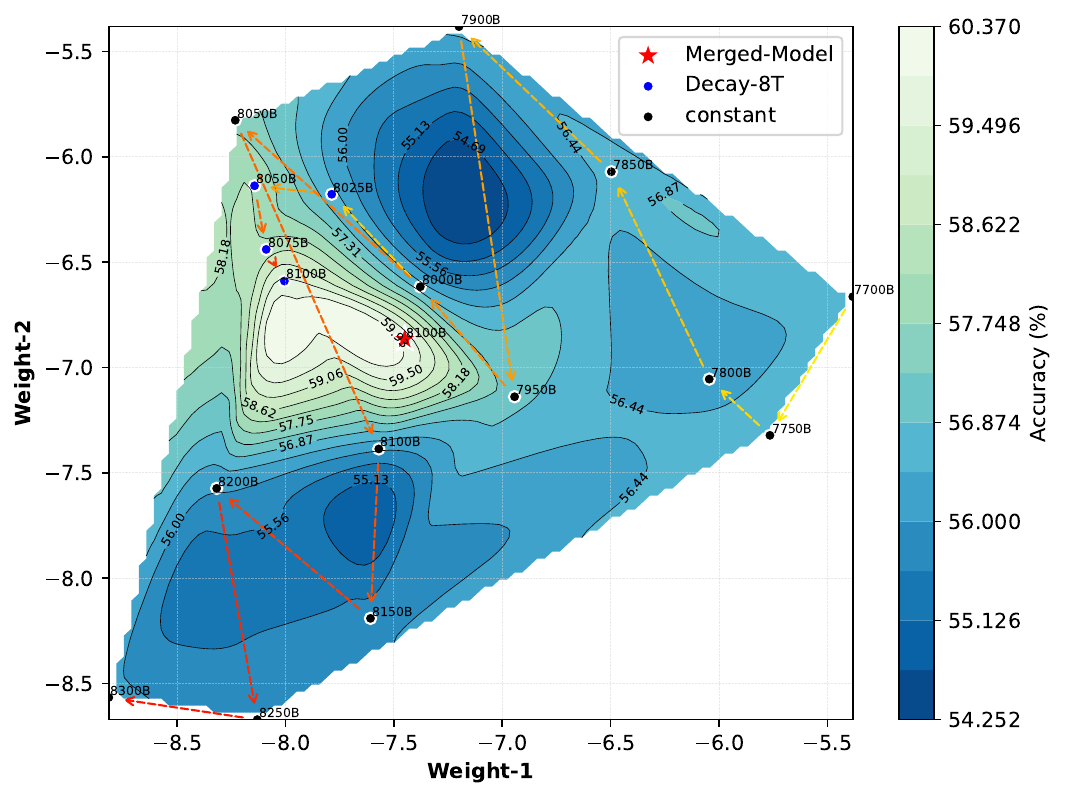}
    \caption{\textbf{Visualization of Model Parameters and Performance Contour Lines.} Points connected by directional arrows represent parameter trajectories during the constant and decay phases, respectively. The red star indicates the solution space of the merged model.}
    \label{fig:wss_ling_mini_merge_ckpt_weight_distributions_for_paper}
\end{figure*}

To visually analyze the correlation between model parameters and performance during training, we employ t-SNE~\citep{vanDerMaaten2008} dimensionality reduction to project the weight matrix of a specific layer into the 2-dimensional embedding space.
This projection is combined with performance evaluations to generate the composite contour map illustrated in Figure \ref{fig:wss_ling_mini_merge_ckpt_weight_distributions_for_paper}.
The directional arrows in the visualization explicitly illustrate the parameter trajectories across both constant and decay phases.
Our experimental analysis revealed two principal findings:
1) During the decay phase, model parameters gradually converge toward the merged model solution space, which achieves superior performance compared to checkpoints with a constant LR.
2) The LR reduction in the decay phase enables more precise parameter refinement than the expansive exploration observed under constant LR. This controlled convergence facilitates localization of nearby optimal solutions. 
We confirm the speculation of previous work~\cite{wen2024understanding}, 
and formalize this dynamic with an analogy: constant LR training resembles traversing a ``canyon'' with oscillating steps, while merging resembles finding a ``river'' at the canyon floor that guides efficient convergence. 

%% file: sections/rela.tex
\section{Related Work}

\subsection{Learning Rate Schedule}
Learning rate (LR) scheduling is critical for training performant models~\citep{Rethinking,Akhilesh}. Classic schedules like Cosine~\citep{kaplan2020scaling,Simple} or Linear~\citep{defazio2023optimal} decay adjust the LR based on a predefined total training duration, which is inflexible for continual training. The Warmup-Stable-Decay (WSD) schedule~\citep{minicpm} addresses this by introducing a stable LR phase after warmup, decoupling the eventual decay from a fixed training length and offering greater flexibility for long or continuous training runs. More recently, researches have explored ``schedule-free'' methods to eliminate the decay phase entirely, which maintain a constant LR and instead leverage weight averaging techniques~\citep{Aaron2024Road,song2025through,Hanlin2025cbs}. Builds upon these schedule-free principles, our work propose to replace WSD's decay phase with a post-hoc checkpoint merging operation instead of specific online averaging strategies. This simplifies the training pipeline and, by formalizing the connection between LR decay and checkpoint merging, allows decay strategy to be theoretically approximated, leading to free offline exploration and enhanced model performance.

\subsection{Model Merging}
Model merging~\citep{Averaging,Modelsoups} has emerged as an efficient paradigm for model construction. This approach achieves effective knowledge transfer and performance improvement through parameter-level integration of multiple models.
Model merging is primarily utilized in two distinct scenarios: (1) The integration of knowledge and capabilities from multiple independently trained models into a single parameter set, with the objective of preserving maximal performance from each specialized model~\citep{commanda,WARP};  and (2) the merging of checkpoints along a single training trajectory, which functions as a form of Polyak averaging~\citep{polyak1992acceleration}. In this application, the merging operation acts as a smoothing mechanism that reduces noise inherent in stochastic gradient-based optimization~\citep{sanyal2023early,Checkpoint,lawa,Trainable,Sandler2023training,1sqrt}. While \cite{pma} show model merging can achieve
performance competitive with decay-based schedule, these techniques have also demonstrated practical utility in industrial-scale LLM development~\citep{grattafiori2024llama,deepseekv3,commanda}.
Our work falls into the second category of merging checkpoints along a single trajectory. We establish a theoretical connection between this operation and learning rate decay and provide a principled approach to convert various
LR decay strategies into a theoretically approximate model averaging implementation.

%% file: sections/conclu.tex
\section{Conclusion}
In this paper, we have presented WSM, a decay-free LR scheduling approach for LLM pre-training. Our method bridges LR decay and checkpoint merging by establishing the theoretical connection. By eliminating the conventional decay phase, WSM simplifies LR scheduling while reformulating various decay strategies as principled model averaging schemes. Through systematic analysis, we identified merge duration as the most critical factor influencing model performance—outweighing other implementation choices. Extensive experiments have demonstrated WSM’s superiority over traditional WSD schedules, with consistent improvements across multiple benchmarks and sustained benefits during fine-tuning. Due to its optimizer-agnostic design, WSM offers broad applicability without modifying the underlying training pipeline. 

As future work, we plan to expand the WSM framework by incorporating additional instantiations of decay strategies, as it can directly convert any decay strategy into a theoretically approximate checkpoint merging scheme. Given its flexibility, we also aim to adapt WSM to more complex tuning scenarios, such as dataset mixture optimization.

%% file: sections/appendix.tex
\appendix

\section{Experimental Settings}
\label{setup}

\paragraph{Model architecture}
The core architectures of our experimental model are detailed in Table~\ref{table:Architectures}. 
The model is configured with 20 layers and a hidden dimension size of 2048. Except for the first layer, all FFNs layers are replaced with MoE layers. We adopt the GQA attention mechanism~\citep{gqa} and integrate Rotary Position Embedding (RoPE)~\citep{rope}, enabling the model to support sequence lengths up to 8K tokens. For parameter initialization, all learnable parameters are randomly initialized using a standard deviation of 0.006. 
Under this configuration, the model consists of a total of 16.3 billion parameters, of which approximately 1.43 billion are activated for each token during inference.

\begin{table}[h]
  \small
  \caption{{\textbf{Detailed model architectures.}}}
  \label{table:Architectures}
  \centering
  \setlength{\tabcolsep}{5pt} %
  \begin{tabular}{cccccccccccc}
    \toprule
      & $n_{layers}$ & $d_{model}$ & $d_{ffn}$ & $d_{expert}$ &$n_{heads}$ & $n_{kv\_head}$ & $E$ & $E_a$ & $E_s$ & $N$ & $N_a$ \\
    \midrule
      & 20 & 2048 & 5120 & 512 & 16 & 4 & 256 & 8 & 1 & 16.3B & 1.43B \\
    \bottomrule
  \end{tabular}
\end{table}

\paragraph{Training hyperparameters}
We use the AdamW optimizer~\citep{adamw} with hyperparameters set as follows: $\beta_1 = 0.9$, $\beta_2 = 0.95$, and weight decay of 0.1. Gradient clipping~\citep{clipping} norm is set to 1.0. According to the scaling laws for MoE optimal hyper-parameters, the maximum learning rates were set to $3.74e{-4}$. The batch size is set to 2048, and with a maximum sequence length of 8K, each training batch contains 16M tokens.

\paragraph{Pre-training dataset}
The training data is sourced from a large-scale multilingual corpus created by the Ling Team, primarily covering English and Chinese, while also including various other languages. This corpus encompasses web text, mathematical materials, programming scripts, published literature, and diverse textual content. To validate model performance, we extracted a 10T-token subset from this corpus for training.

\paragraph{Evaluation setup}
\label{}
To evaluate performance, we consider a diverse suite of downstream tasks designed to provide a holistic assessment of model capabilities. For base model, tasks are grouped into several categories, such as: (a) General Knowledge/Reasoning (e.g., ARC~\citep{arc}, AGIEval~\citep{agieval}, OpenBookQA~\citep{openbookqa}, BBH~\citep{bbh}, WorldSense~\citep{worldsense}, PIQA~\citep{piqa}, hellaswag~\citep{hellaswag} and KOR-Bench~\citep{korbench}) (b) Language Understanding (e.g., race~\citep{race}, SQuAD 2.0~\citep{squad2}, TriviaQA~\citep{trivalqa}, NQ~\citep{nq} and winogrande~\citep{winogrande}) (c) Professional Knowledge (e.g., MMLU~\citep{mmlu}, CMMLU~\citep{cmmlu}, C-Eval~\citep{ceval}, MMLU-Pro~\citep{mmlu-pro}, GPQA~\citep{gpqa} and SuperGPQA~\citep{supergpqa}) (d) Math (e.g., GSM8K~\citep{gsm8k}, MATH~\citep{math}, gaokao~\citep{gaokao}, GSM-Plus~\citep{gsm-plus}, mgsm-zh~\citep{mgsm}, CMATH~\citep{cmath}, MathBench~\citep{mathbench}, minerva\_math~\citep{math}, college\_math~\citep{college-math} and cn\_middle\_school\_24) (e) Code (e.g., HumanEval~\citep{humaneval}, LiveCodeBench~\citep{livecodebench}, MBPP~\citep{mbpp}, HumanEval\_plus~\citep{mbpp+}, MBPP\_plus~\citep{mbpp+}, HumanEval\_cn~\citep{humaneval_cn}, HumanEval\_fim~\citep{fim} and CruxEval~\citep{cruxeval}). For SFT model, the categories and tasks are shown in Table \ref{inst_dataset_1} and \ref{inst_dataset_2}.

\section{Details of Theoretical Connection Between Decay and Merging}
\label{app:Theoretical}
The core idea of checkpoint merging in this work is to take an ordered list of checkpoints, $[\theta_n, \theta_{n+1}, \dots, \theta_{n+k}]$, and apply a merge function to generate a single model $\hat{\theta}_{n+k}$. Here, $\theta_i \in \mathbb{R}^d$ represents the model's parameter vector at the $i$-th training iteration.

The most general form is a weighted average of the checkpoints:
\begin{equation}
    \hat{\theta}_{n+k} = \sum_{j=0}^{k} c_j \theta_{n+j}
    \label{eq:weighted_avg_appendix}
\end{equation}
where $\{c_j\}$ are non-negative weights that sum to one, i.e., $\sum_{j=0}^{k} c_j = 1$.

While intuitive, this formulation obscures a deeper connection to the training dynamics. We can reveal this link by expressing each checkpoint in terms of an initial checkpoint $\theta_n$ and the subsequent gradient updates.

We assume that the updates between checkpoints at different time steps are independent. Let $g_{i}$ be the gradient update vector (including the learning rate) at step $i$, such that the model is updated via $\theta_{i+1} = \theta_i - g_i$. Any checkpoint $\theta_{n+j}$ can therefore be written as the sum of an initial state $\theta_n$ and the sequence of negative gradient updates that followed:
\begin{equation}
    \theta_{n+j} = \theta_{n} - \sum_{l=1}^{j} g_{n+l-1}
    \label{eq:theta_as_gradient_sum_appendix}
\end{equation}

Substituting Eq. \ref{eq:theta_as_gradient_sum_appendix} into the general merging formula (Eq. \ref{eq:weighted_avg_appendix}) yields:
\begin{align}
    \hat{\theta}_{n+k} &= \sum_{j=0}^{k} c_j \left( \theta_n - \sum_{l=1}^{j} g_{n+l-1} \right) \\
    &= \left( \sum_{j=0}^{k} c_j \right) \theta_n - \sum_{j=0}^{k} c_j \sum_{l=1}^{j} g_{n+l-1} \\
    &= \theta_n - \sum_{j=1}^{k} c_j \sum_{l=1}^{j} g_{n+l-1} \label{eq:rearrange_sum_step1_app}
\end{align}

The double summation in Eq. \ref{eq:rearrange_sum_step1_app} can be rearranged by changing the order of summation. A gradient update $g_{n+i-1}$ is included in the sum for all checkpoints $\theta_{n+j}$ where $j \ge i$.
\begin{align}
    \sum_{j=1}^{k} c_j \sum_{l=1}^{j} g_{n+l-1} &= \sum_{i=1}^{k} \left( \sum_{j=i}^{k} c_j \right) g_{n+i-1}
\end{align}
This shows that a weight coefficient of $\sum_{j=i}^{k} c_j$ is applied to the gradient update $g_{n+i-1}$ from step $n+i-1$.
By defining a new set of weights for the gradient updates, $w_i = \sum_{j=i}^{k} c_j$, we arrive at the final equivalent expression for checkpoint merging:
\begin{equation}
    \hat{\theta}_{n+k} = \theta_n - \sum_{i=1}^{k} w_i \cdot g_{n+i-1}
    \label{eq:merge_as_gradient_reweighting_app}
\end{equation}
This equation demonstrates that merging checkpoints with weights $\{c_j\}$ is equivalent to applying a synthetic decay schedule defined by weights $\{w_i\}$ to the gradients accumulated after the base checkpoint $\theta_n$, and there exists a mapping between the merging weights ${c_j}$ and the effective learning rates ${w_i}$
\subsection{Proof of Theorem \ref{thm:inverse}}

We seek to find the unique checkpoint weights $\{c_j\}$ corresponding to a given desired sequence of gradient decay coefficients $\{w_i\}_{i=1}^k$. The relationship derived in the previous section is the starting point:
\begin{equation}
w_i = \sum_{j=i}^{k} c_j
\label{eq:w_from_c_proof}
\end{equation}

We can solve for the checkpoint weights $\{c_j\}$ by starting from the last element and working backwards.

For $i = k$, the sum in Eq. \ref{eq:w_from_c_proof} has only one term:
\begin{equation}
w_k = \sum_{j=k}^{k} c_j = c_k
\label{eq:proof_ck}
\end{equation}
This gives us the value of $c_k$ directly.

For any $i \in [1, k-1]$, we can write out the expressions for $w_i$ and $w_{i+1}$:
\begin{align*}
w_i &= c_i + c_{i+1} + c_{i+2} + \dots + c_k \\
w_{i+1} &= \qquad c_{i+1} + c_{i+2} + \dots + c_k
\end{align*}
Subtracting the second equation from the first yields the expression for $c_i$:
\begin{equation}
w_i - w_{i+1} = c_i
\label{eq:proof_cj}
\end{equation}

Finally, for $c_0$, we use the constraint that the checkpoint weights must sum to one: $\sum_{j=0}^{k} c_j = 1$.
\begin{align}
c_0 &= 1 - \sum_{j=1}^{k} c_j \nonumber \\
&= 1 - \left( c_1 + c_2 + \dots + c_{k-1} + c_k \right) \nonumber \\
&= 1 - \left( (w_1 - w_2) + (w_2 - w_3) + \dots + (w_{k-1} - w_k) + w_k \right) \label{eq:telescoping_sum}
\end{align}
The sum in the parentheses is a telescoping series which simplifies to $w_1$.
\begin{equation}
c_0 = 1 - w_1
\label{eq:proof_c0}
\end{equation}

This completes the derivation of the unique formulas for $\{c_j\}$ as stated in the theorem.

For the checkpoint weights $\{c_j\}$ to be valid, they must be non-negative. This imposes conditions on the sequence $\{w_i\}$.
\begin{itemize}[leftmargin=*]
    \item From Eq. \ref{eq:proof_ck}, $c_k \ge 0$ implies $w_k \ge 0$.
    \item From Eq. \ref{eq:proof_cj}, $c_j \ge 0$ for $j \in [1, k-1]$ implies $w_j - w_{j+1} \ge 0$, which means $w_j \ge w_{j+1}$. This shows that the sequence $\{w_i\}$ must be monotonically non-increasing.
    \item From Eq. \ref{eq:proof_c0}, $c_0 \ge 0$ implies $1 - w_1 \ge 0$, which means $w_1 \le 1$.
\end{itemize}
Combining these conditions, we arrive at the requirement that the gradient decay sequence must be bounded and monotonically non-increasing: $1 \ge w_1 \ge w_2 \ge \dots \ge w_k \ge 0$. This ensures that a valid (non-negative) set of checkpoint weights $\{c_j\}$ can be derived.

\section{Additional Experiments}
\label{app:add_exp}

\begin{figure*}[h]
    \centering
    \includegraphics[width=\textwidth]{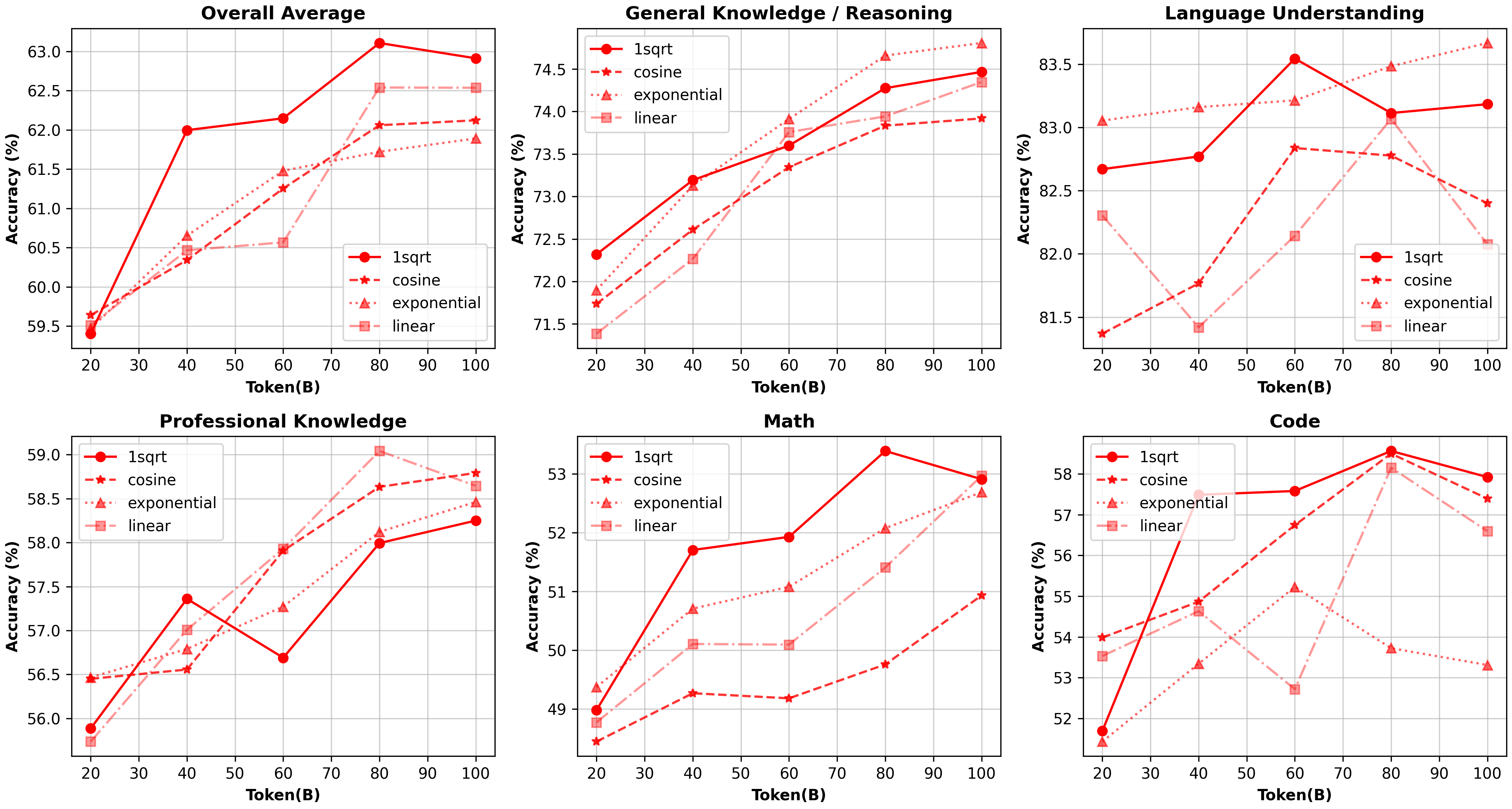}
    \caption{\textbf{Comprehensive performance comparison between different decay strategy of WSM schedule.}}
    \label{fig:app_decay}
\end{figure*}

Our preliminary experiments into Warmup-Stable-Decay (WSD) learning rate schedules revealed a clear performance hierarchy among different decay curves, with the 1-sqrt strategy emerging as superior. 
Specifically, we conducted a controlled experiment initialized from a Ling-lite~\citep{team2025every} base checkpoint that was pre-trained on 7T tokens. We then annealed the model for an additional 100B tokens using a consistent, high-quality dataset, where we varied only the annealing decay function. The final learning rate for all experimental runs was set to zero. The results, depicted in Figure~\ref{fig:app_decay}, confirm that the 1-sqrt decay outperforms other methods in benchmarks. Based on this evidence, we establish the WSD schedule with 1-sqrt decay as a strong baseline for all subsequent experiments. As this was a preliminary study, the starting checkpoint used here differs from that used in our main experiments in Section~\ref{sec:exp}.

\section{How Checkpoint Merging Influences Network Behavior: Stability \& Generalizability}
\label{parameter}
\begin{figure*}[t]
    \centering
    \includegraphics[width=\textwidth]{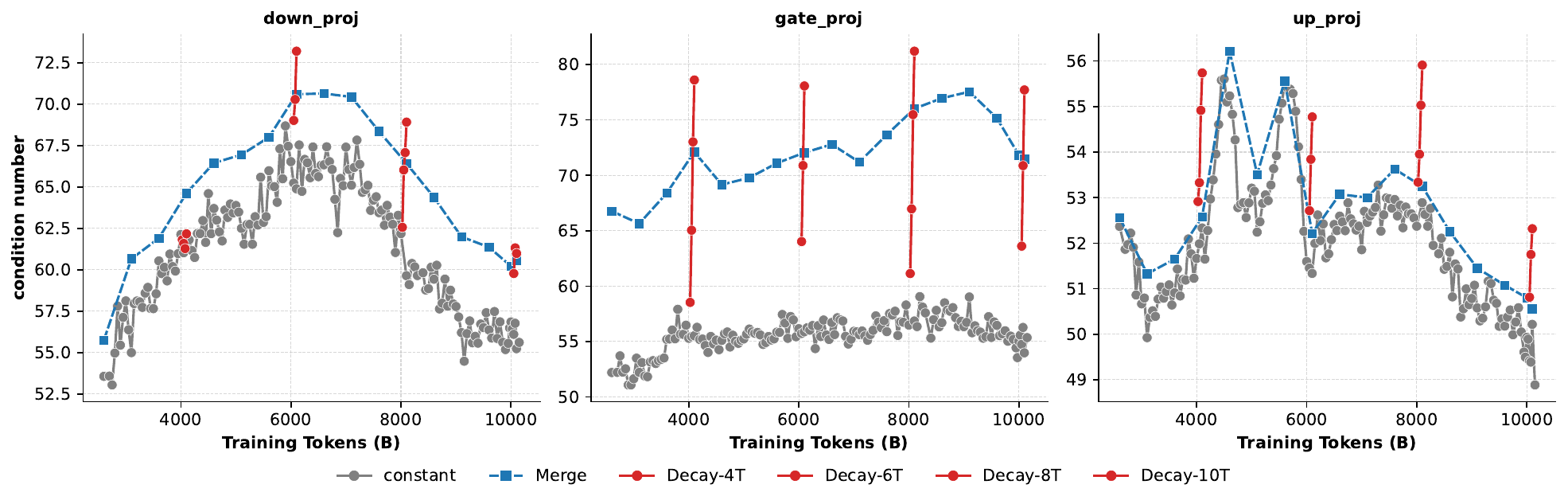}
    \caption{\textbf{Condition number of weight matrices across different training tokens.}}
    \label{fig:wss_ling_mini_merge_ckpt_condition_number}
\end{figure*}

The architecture composed of sequentially connected transformer blocks has been widely adopted in mainstream LLMs. In such chain-like structures, variations in shallow-layer outputs typically propagate layer-wise. More stable and generalizable input-output patterns exhibit greater potential for enhancing downstream task performance. To investigate how checkpoint merging influences network behavior, we conduct the following analyses:

\paragraph{Stability}
In numerical analysis, the condition number quantifies a function’s sensitivity to input perturbations and the resultant output errors. Remarkably, our analysis in Figure \ref{fig:wss_ling_mini_merge_ckpt_condition_number} reveals that the decay process induces sharp deterioration of condition numbers, whereas checkpoint merging demonstrates superior stability. This indicates that the checkpoint merging strategy not only improves performance but also preserves parametric stability.

\begin{figure*}[t]
    \centering
    \includegraphics[width=\textwidth]{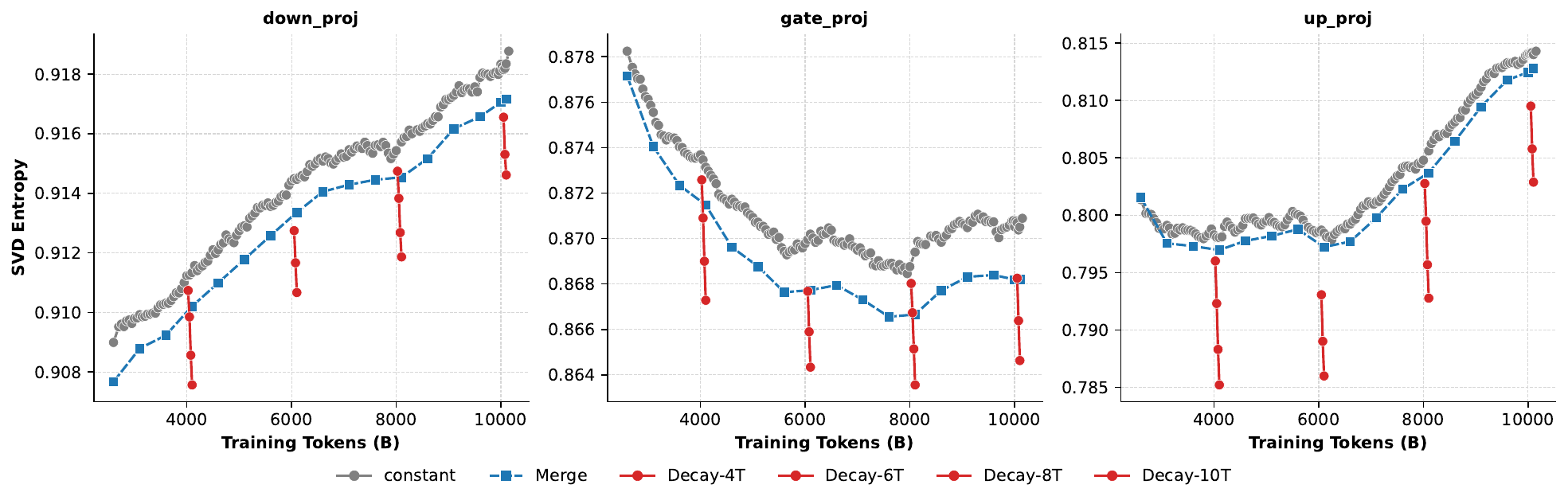}
    \caption{\textbf{The entropy of singular values for each weight matrix in the feed-forward network (FFN) layers.}}
    \label{fig:wss_ling_mini_merge_ckpt_svd_entropy}
\end{figure*}

\paragraph{Generalizability}
When maintaining competitive performance, higher SVD entropy (singular values entropy) correlates with greater matrix effective rank, indicating greater information capacity in matrix operations. For continue pre-training and fine-tuning scenarios, higher SVD Entropy often means higher potential for model plasticity~\citep{alter2000singular,roy2007effective,liu2025muonscalablellmtraining}. Figure \ref{fig:wss_ling_mini_merge_ckpt_svd_entropy} shows the trend of SVD Entropy during training. We observe that decay is a rapid entropy reduction process, which continuously damages the potential for future continue training of the model. In contrast, the merged models still maintain a higher generalizability, manifested as a higher SVD Entropy.

\section{Detailed Evaluation Results}
\begin{figure*}[ht]
    \centering
    \includegraphics[width=0.95\textwidth]{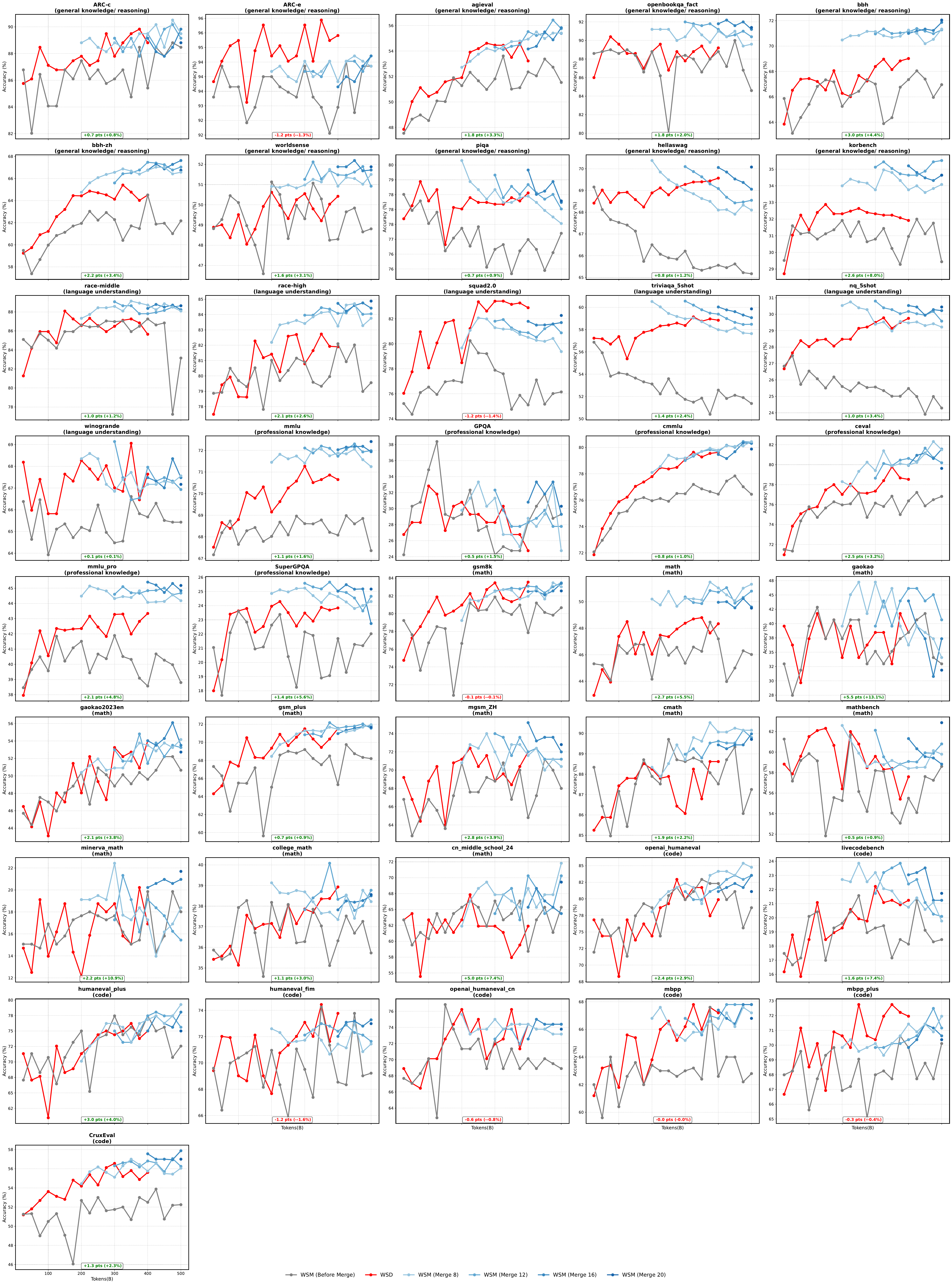}
    \caption{Detailed performance comparison between the standard WSD schedule (via learning rate decay) and our WSM schedule (via checkpoint merging) under different merging durations.}
    \label{fig:all_datasets}
\end{figure*}

We provide detailed evaluation results to compare our methods with WSD. Figure \ref{fig:all_datasets} shows a detailed comparison on each dataset across the various categories from the main experiments in Section \ref{sec:main}, where WSM achieves an advantage on the vast majority of datasets. We select the checkpoint with the highest average score for each method, including WSD and the three WSM merging algorithms, and list them in Table \ref{table:sample-table}. Table \ref{inst_dataset_1} and \ref{inst_dataset_2} shows the performance comparison of checkpoints generated by the WSM and WSD schedules after supervised fine-tuning (SFT).
\label{all_datasets}

\begin{table}[h]
\small
  \caption{\textbf{Detailed performance comparison of base models trained using WSM (with three distinct merging algorithms) versus WSD scheduling approaches.}}
  \label{table:sample-table}
  \centering
  \setlength{\tabcolsep}{5pt} %
  \begin{tabular}{clllllll}
\toprule
\multicolumn{1}{l}{}                           & \multicolumn{1}{l}{\multirow{2}{*}{\textbf{Metric}}} & & \multicolumn{1}{c}{\multirow{2}{*}{\textbf{WSD}}} &  & \multicolumn{3}{c}{\textbf{WSM}} \\ \cmidrule(l){5-8} 
\multicolumn{1}{l}{}                           & \multicolumn{1}{c}{} & \multicolumn{1}{c}{}               &      &  & \textbf{EMA}    & \textbf{mean}  & \textbf{1-\textit{sqrt}}  \\
\midrule
\multirow{10}{*}{\shortstack{General Knowledge \\ \& Reasoning}} & ARC-c                  &  & 88.8                                     &  & 87.8   & 88.1  & 89.8   \\
                                               & ARC-e                  &  & 95.4                                     &  & 93.7   & 94.0  & 94.5   \\
                                               & AGIEval                &  & 53.2                                     &  & 53.8   & 54.4  & 54.2   \\
                                               & OpenBookQA\_fact       &  & 89.2                                     &  & 89.2   & 92.2  & 91.8   \\
                                               & BBH                    &  & 69.0                                     &  & 70.0   & 71.2  & 71.0   \\
                                               & BBH-zh                 &  & 64.5                                     &  & 65.3   & 67.3  & 66.5   \\
                                               & WorldSense             &  & 50.4                                     &  & 50.6   & 51.9  & 52.1   \\
                                               & PIQA                   &  & 78.6                                     &  & 78.2   & 78.5  & 78.5   \\
                                               & HellaSwag              &  & 69.6                                     &  & 68.2   & 69.9  & 69.7   \\
                                               & KOR-Bench               &  & 31.9                                     &  & 33.7   & 34.8  & 34.6   \\
\midrule
\multirow{8}{*}{\shortstack{Language \\ Understanding}}        & race-middle            &  & 85.7                                     &  & 88.7   & 88.8  & 87.9   \\
                                               & race-high              &  & 81.9                                     &  & 82.2   & 84.2  & 83.9   \\
                                               & SQuAD2.0               &  & 82.9                                     &  & 81.6   & 81.5  & 81.3   \\
                                               & TriviaQA\_5shot        &  & 58.8                                     &  & 57.5   & 59.8  & 59.7   \\
                                               & NQ\_5shot              &  & 29.8                                     &  & 29.1   & 30.4  & 30.3   \\
                                               & WinoGrande             &  & 67.6                                     &  & 66.7   & 67.3  & 66.6   \\
\midrule
\multirow{6}{*}{\shortstack{Professional \\ Knowledge}}        & MMLU                   &  & 70.7                                     &  & 71.2   & 72.2  & 72.1   \\
                                               & GPQA                   &  & 24.8                                     &  & 30.8   & 33.3  & 30.8   \\
                                               & CMMLU                  &  & 79.6                                     &  & 78.3   & 79.2  & 79.3   \\
                                               & MMLU-Pro              &  & 43.3                                     &  & 43.8   & 45.2  & 44.7   \\
                                               & SuperGPQA              &  & 23.8                                     &  & 24.3   & 25.5  & 25.7   \\
                                               & C-Eval                  &  & 78.5                                     &  & 78.3   & 80.9  & 79.9   \\
\midrule
\multirow{11}{*}{\shortstack{Math}}                         & gsm8k                  &  & 83.6                                     &  & 82.0   & 82.6  & 82.3   \\
                                               & MATH                   &  & 48.3                                     &  & 50.8   & 50.0  & 49.6   \\
                                               & gaokao                 &  & 38.5                                     &  & 47.3   & 39.6  & 48.4   \\
                                               & gaokao2023en           &  & 54.0                                     &  & 49.1   & 53.5  & 53.0   \\
                                               & gsm\_plus              &  & 71.5                                     &  & 69.3   & 71.4  & 70.9   \\
                                               & mgsm\_zh               &  & 72.0                                     &  & 73.2   & 73.2  & 72.8   \\
                                               & CMATH                  &  & 88.6                                     &  & 88.4   & 89.3  & 89.2   \\
                                               & MathBench              &  & 57.6                                     &  & 61.8   & 60.3  & 61.6   \\
                                               & minerva\_math          &  & 16.9                                     &  & 19.9   & 20.6  & 19.1   \\
                                               & college\_math          &  & 38.9                                     &  & 37.8   & 38.3  & 39.0   \\
                                               & cn\_middle\_school\_24 &  & 62.4                                     &  & 67.3   & 68.3  & 70.3   \\
\midrule
\multirow{8}{*}{\shortstack{Code}}                          & HumanEval      &  & 79.9                                     &  & 80.5   & 81.7  & 81.1   \\
                                               & LiveCodeBench          &  & 21.2                                     &  & 20.3   & 23.2  & 22.7   \\
                                               & HumanEval\_plus        &  & 75.0                                     &  & 75.6   & 77.4  & 77.4   \\
                                               & HumanEval\_fim         &  & 73.8                                     &  & 69.3   & 73.1  & 73.9   \\
                                               & HumanEval\_cn  &  & 74.4                                     &  & 77.4   & 75.0  & 75.6   \\
                                               & MBPP                   &  & 67.2                                     &  & 65.0   & 66.8  & 66.8   \\
                                               & MBPP\_plus             &  & 72.0                                     &  & 70.6   & 70.4  & 70.6   \\
                                               & CruxEval               &  & 55.6                                     &  & 54.8   & 57.0  & 57.4  \\
\bottomrule
\end{tabular}
\end{table}

\begin{table}[h]
\small
  \caption{Detailed performance comparison of checkpoints generated by the WSM and WSD schedule after supervised fine-tuning (SFT). Both base checkpoints are fine-tuned under identical settings for 5 epochs. Results are reported based on the epoch with the highest average benchmark score.}
  \label{inst_dataset_1}
  \centering
  \setlength{\tabcolsep}{5pt} %
\begin{tabular}{ccccll}
\toprule
\textbf{}                            & \textbf{}                                                                                       & \textbf{Metric}            &  & \textbf{WSD}   & \textbf{WSM}   \\ \midrule

\multirow{14}{*}{{Knowledge}} & \multirow{6}{*}{{\begin{tabular}[c]{@{}c@{}}Basic\\      Knowledge\end{tabular}}}        & ARC-c               &  & 90.85 & 89.49 \\
                                     &                                                                                                 & BoolQ               &  & 84.80  & 85.38 \\
                                     &                                                                                                 & GaokaoBench         &  & 75.70  & 79.95 \\
                                     &                                                                                                 & AGIEval             &  & 61.87 & 65.22 \\
                                     &                                                                                                 & NQ                  &  & 25.4  & 26.43 \\
                                     &                                                                                                 & TriviaQA            &  & 53.93 & 55.52 \\ \cmidrule{2-6} 
                                     & {Average}                                                                                &                     &  & 65.42 & 67.00    \\ \cmidrule{2-6} 
                                     & \multirow{6}{*}{{\begin{tabular}[c]{@{}c@{}}Professional\\      Knowledge\end{tabular}}} & C-Eval               &  & 76.37 & 77.87 \\
                                     &                                                                                                 & CMMLU               &  & 76.13 & 76.78 \\
                                     &                                                                                                 & MMLU                &  & 72.76 & 74.59 \\
                                     &                                                                                                 & MMLU-Pro           &  & 46.09 & 49.67 \\
                                     &                                                                                                 & GPQA                &  & 29.55 & 33.4  \\
                                     &                                                                                                 & SuperGQPA           &  & 26.50  & 26.43 \\ \cmidrule{2-6} 
                                     & {Average}                                                                                &                     &  & 54.57 & 56.46 \\ \midrule
\multirow{18}{*}{{Code}}      & \multirow{6}{*}{{Code Completion}}                                                       & HumanEval           &  & 85.90  & 86.20  \\
                                     &                                                                                                 & HumanEval\_plus     &  & 81.10  & 80.95 \\
                                     &                                                                                                 & HumanEval\_cn       &  & 78.20  & 76.07 \\
                                     &                                                                                                 & CruxEval            &  & 59.69 & 58.88 \\
                                     &                                                                                                 & Multiple            &  & 66.14 & 64.75 \\
                                     &                                                                                                 & HumanEvalFix        &  & 63.01 & 62.50  \\ \cmidrule{2-6} 
                                     & {Average}                                                                                &                     &  & 72.34 & 71.56 \\ \cmidrule{2-6} 
                                     & \multirow{6}{*}{{Code Generation}}                                                      & MBPP                &  & 81.03 & 81.97 \\ 
                                     &                                                                                                 & MBPP\_plus          &  & 73.02 & 71.43 \\
                                     &                                                                                                 & LiveCodeBench       &  & 31.31 & 33.31 \\
                                     &                                                                                                 & BigCodeBench        &  & 33.77 & 33.16 \\
                                     &                                                                                                 & CodeForces          &  & 19.60  & 18.07 \\
                                     &                                                                                                 & Bird-SQL            &  & 25.95 & 28.10  \\ \cmidrule{2-6} 
                                     & {Average}                                                                                &                     &  & 44.11 & 44.34  \\ \midrule
\multirow{18}{*}{{Math}}      & \multirow{5}{*}{{Elementary Mathematics}}                                                & CMATH               &  & 93.08 & 93.99 \\
                                     &                                                                                                 & gsm8k               &  & 87.34 & 87.79 \\
                                     &                                                                                                 & cn\_middle\_school\_24 &  & 72.28 & 73.27 \\
                                     &                                                                                                 & mgsm\_zh            &  & 76.80  & 78.25 \\
                                     &                                                                                                 & gsm\_plus           &  & 77.28 & 77.93 \\ \cmidrule{2-6} 
                                     & {Average}                                                                                &                     &  & 81.36 & 82.25 \\ \cmidrule{2-6} 
                                     & \multirow{7}{*}{{Intermediate Mathematics}}                                              & MATH                &  & 74.24 & 77.08 \\
                                     &                                                                                                 & MathBench           &  & 74.72 & 75.74 \\
                                     &                                                                                                 & college math        &  & 43.57 & 43.52 \\
                                     &                                                                                                 & gaokao              &  & 64.17 & 60.77 \\
                                     &                                                                                                 & minerva math        &  & 55.74 & 57.31 \\
                                     &                                                                                                 & gaokao2023en        &  & 63.12 & 65.19 \\
                                     &                                                                                                 & MATH500             &  & 74.75 & 77.05 \\ \cmidrule{2-6} 
                                     & {Average}                                                                                &                     &  & 64.33 & 65.24 \\ \cmidrule{2-6} 
                                     & \multirow{3}{*}{{Advanced Mathematics}}                                                  & OlympiadBench       &  & 41.89 & 41.93 \\
                                     &                                                                                                 & AIME2024            &  & 11.04 & 11.67 \\
                                     &                                                                                                 & AIME2025            &  & 11.46 & 12.71 \\ \cmidrule{2-6} 
                                     & {Average}                                                                                &                     &  & 21.46 & 22.10  \\ 
\bottomrule
\end{tabular}
\end{table}

\begin{table}[h]
\small
  \caption{\textbf{Detailed performance comparison of checkpoints generated by the WSM and WSD schedule after supervised fine-tuning (SFT).} Both base checkpoints are fine-tuned under identical settings for 5 epochs. Results are reported based on the epoch with the highest average benchmark score.}
  \label{inst_dataset_2}
  \centering
  \setlength{\tabcolsep}{5pt} %
\begin{tabular}{ccccll}
\toprule
\textbf{}                            & \textbf{}                                                                                       & \textbf{Metric}            &  & \textbf{WSD}   & \textbf{WSM}   \\ \midrule
\multirow{5}{*}{{Language}}   & \multirow{4}{*}{{Language Understanding}}                                                & C3                  &  & 84.38 & 87.18 \\
                                     &                                                                                                 & WSC                 &  &  69.23     & 77.88 \\
                                     &                                                                                                 & race-high           &  & 82.93 & 84.65 \\
                                     &                                                                                                 & race-middle         &  & 87.95 & 89.42 \\ \cmidrule{2-6} 
                                     & {Average}                                                                                &                     &  & 81.12 & 84.78 \\ \midrule
\multirow{11}{*}{{Reasoning}} & \multirow{10}{*}{{Complex Reasoning}}                                                    & bbh                 &  & 72.87 & 73.97 \\
                                     &                                                                                                 & drop                &  & 76.41 & 78.66 \\
                                     &                                                                                                 & hellaswag           &  & 68.11 & 70.93 \\
                                     &                                                                                                 & ocnli               &  & 51.32 & 52.75 \\
                                     &                                                                                                 & piqa                &  & 81.50  & 82.92 \\
                                     &                                                                                                 & ProntoQA            &  & 37.00    & 38.00    \\
                                     &                                                                                                 & Multi-LogiEval      &  & 0.27  & 56.68 \\
                                     &                                                                                                 & MuSR                &  & 48.53 & 51.33 \\
                                     &                                                                                                 & korbench            &  & 37.52 & 37.52 \\
                                     &                                                                                                 & bbh-zh              &  & 69.38 & 71.47 \\ \cmidrule{2-6} 
                                     & {Average}                                                                                &                     &  & 63.21 & 64.94 \\ \midrule
\multirow{9}{*}{{Agent}}      & \multirow{5}{*}{{Tool-use}}                                                              & teval\_v2\_en       &  & 84.4  & 86.16 \\
                                     &                                                                                                 & teval\_v2\_zh       &  & 83.49 & 84.72 \\
                                     &                                                                                                 & BFCL\_AST           &  & 79.35 & 78.30  \\
                                     &                                                                                                 & BFCL-Live           &  & 68.33 & 71.39 \\
                                     &                                                                                                 & NEXUS               &  & 30.51 & 29.21 \\ \cmidrule{2-6} 
                                     & {Average}                                                                                &                     &  & 69.22 & 69.96 \\ \cmidrule{2-6} 
                                     & \multirow{2}{*}{{Instruction Following}}                                                 & IFEval              &  & 71.9  & 75.71 \\ 
                                     &                                                                                                 & alignbench          &  & 59.10  & 59.80  \\ \cmidrule{2-6} 
                                     & {Average}                                                                                &                     &  & 65.5  & 67.75 \\ 
\bottomrule
\end{tabular}
\end{table}

%% file: main.bbl
\begin{thebibliography}{70}
\providecommand{\natexlab}[1]{#1}
\providecommand{\url}[1]{\texttt{#1}}
\expandafter\ifx\csname urlstyle\endcsname\relax
  \providecommand{\doi}[1]{doi: #1}\else
  \providecommand{\doi}{doi: \begingroup \urlstyle{rm}\Url}\fi

\bibitem[Aakanksha et~al.(2025)Aakanksha, Ahmadian, Ahmed, Alammar, Alizadeh, Alnumay, Althammer, Arkhangorodsky, Aryabumi, Aumiller, Avalos, Aviv, Bae, Baji, Barbet, Bartolo, Bebensee, Beladia, Beller{-}Morales, B{\'{e}}rard, Berneshawi, Bialas, Blunsom, Bobkin, Bongale, Braun, Brunet, Cahyawijaya, Cairuz, Campos, Cao, Cao, Castagn{\'{e}}, Cendrero, Currie, Chandak, Chang, Chatziveroglou, Chen, Cheng, Chevalier, Chiu, Choi, Choi, Chung, Cirik, Cismaru, Clavier, Conklin, Crawhall{-}Stein, Crouse, Cruz{-}Salinas, Cyrus, D'souza, Dalla{-}Torre, Dang, Darling, Domingues, Dash, Debugne, Dehaze, Desai, Devassy, Dholakia, Duffy, Edalati, Eldeib, Elkady, Elsharkawy, Erg{\"{u}}n, Ermis, Fadaee, Fan, Fayoux, Flet{-}Berliac, Frosst, Gall{\'{e}}, Galuba, Garg, Geist, Azar, Gilsenan{-}McMahon, Goldfarb{-}Tarrant, Goldsack, Gomez, Gonzaga, Govindarajan, Govindassamy, Grinsztajn, Gritsch, Gu, Guo, Haefeli, Hajjar, Hawes, He, Hofst{\"{a}}tter, and Hong]{commanda}
Aakanksha, Arash Ahmadian, Marwan Ahmed, Jay Alammar, Milad Alizadeh, Yazeed Alnumay, Sophia Althammer, Arkady Arkhangorodsky, Viraat Aryabumi, Dennis Aumiller, Rapha{\"{e}}l Avalos, Zahara Aviv, Sammie Bae, Saurabh Baji, Alexandre Barbet, Max Bartolo, Bj{\"{o}}rn Bebensee, Neeral Beladia, Walter Beller{-}Morales, Alexandre B{\'{e}}rard, Andrew Berneshawi, Anna Bialas, Phil Blunsom, Matt Bobkin, Adi Bongale, Sam Braun, Maxime Brunet, Samuel Cahyawijaya, David Cairuz, Jon~Ander Campos, Cassie Cao, Kris Cao, Roman Castagn{\'{e}}, Juli{\'{a}}n Cendrero, Leila~Chan Currie, Yash Chandak, Diane Chang, Giannis Chatziveroglou, Hongyu Chen, Claire Cheng, Alexis Chevalier, Justin~T. Chiu, Eugene Choi, Eujeong Choi, Tim Chung, Volkan Cirik, Ana Cismaru, Pierre Clavier, Henry Conklin, Lucas Crawhall{-}Stein, Devon Crouse, Felipe Cruz{-}Salinas, Ben Cyrus, Daniel D'souza, Hugo Dalla{-}Torre, John Dang, William Darling, Omar~Darwiche Domingues, Saurabh Dash, Antoine Debugne, Th{\'{e}}o Dehaze, Shaan Desai, Joan Devassy,
  Rishit Dholakia, Kyle Duffy, Ali Edalati, Ace Eldeib, Abdullah Elkady, Sarah Elsharkawy, Irem Erg{\"{u}}n, Beyza Ermis, Marzieh Fadaee, Boyu Fan, Lucas Fayoux, Yannis Flet{-}Berliac, Nick Frosst, Matthias Gall{\'{e}}, Wojciech Galuba, Utsav Garg, Matthieu Geist, Mohammad~Gheshlaghi Azar, Ellen Gilsenan{-}McMahon, Seraphina Goldfarb{-}Tarrant, Tomas Goldsack, Aidan~N. Gomez, Victor~Machado Gonzaga, Nithya Govindarajan, Manoj Govindassamy, Nathan Grinsztajn, Nikolas Gritsch, Patrick Gu, Shangmin Guo, Kilian Haefeli, Rod Hajjar, Tim Hawes, Jingyi He, Sebastian Hofst{\"{a}}tter, and Sungjin Hong.
\newblock Command {A:} an enterprise-ready large language model.
\newblock \emph{CoRR}, abs/2504.00698, 2025.
\newblock \doi{10.48550/ARXIV.2504.00698}.
\newblock \url{https://doi.org/10.48550/arXiv.2504.00698}.

\bibitem[Ainslie et~al.(2023)Ainslie, Lee{-}Thorp, de~Jong, Zemlyanskiy, Lebr{\'{o}}n, and Sanghai]{gqa}
Joshua Ainslie, James Lee{-}Thorp, Michiel de~Jong, Yury Zemlyanskiy, Federico Lebr{\'{o}}n, and Sumit Sanghai.
\newblock {GQA:} training generalized multi-query transformer models from multi-head checkpoints.
\newblock In Houda Bouamor, Juan Pino, and Kalika Bali, editors, \emph{Proceedings of the 2023 Conference on Empirical Methods in Natural Language Processing, {EMNLP} 2023, Singapore, December 6-10, 2023}, pages 4895--4901. Association for Computational Linguistics, 2023.
\newblock \doi{10.18653/V1/2023.EMNLP-MAIN.298}.
\newblock \url{https://doi.org/10.18653/v1/2023.emnlp-main.298}.

\bibitem[Alter et~al.(2000)Alter, Brown, and Botstein]{alter2000singular}
Orly Alter, Patrick~O Brown, and David Botstein.
\newblock Singular value decomposition for genome-wide expression data processing and modeling.
\newblock \emph{Proceedings of the National Academy of Sciences}, 97\penalty0 (18):\penalty0 10101--10106, 2000.

\bibitem[Bavarian et~al.(2022)Bavarian, Jun, Tezak, Schulman, McLeavey, Tworek, and Chen]{fim}
Mohammad Bavarian, Heewoo Jun, Nikolas Tezak, John Schulman, Christine McLeavey, Jerry Tworek, and Mark Chen.
\newblock Efficient training of language models to fill in the middle.
\newblock \emph{CoRR}, abs/2207.14255, 2022.
\newblock \doi{10.48550/ARXIV.2207.14255}.
\newblock \url{https://doi.org/10.48550/arXiv.2207.14255}.

\bibitem[Bhakthavatsalam et~al.(2021)Bhakthavatsalam, Khashabi, Khot, Mishra, Richardson, Sabharwal, Schoenick, Tafjord, and Clark]{arc}
Sumithra Bhakthavatsalam, Daniel Khashabi, Tushar Khot, Bhavana~Dalvi Mishra, Kyle Richardson, Ashish Sabharwal, Carissa Schoenick, Oyvind Tafjord, and Peter Clark.
\newblock Think you have solved direct-answer question answering? try arc-da, the direct-answer {AI2} reasoning challenge.
\newblock \emph{CoRR}, abs/2102.03315, 2021.
\newblock \url{https://arxiv.org/abs/2102.03315}.

\bibitem[Bisk et~al.(2020)Bisk, Zellers, Bras, Gao, and Choi]{piqa}
Yonatan Bisk, Rowan Zellers, Ronan~Le Bras, Jianfeng Gao, and Yejin Choi.
\newblock {PIQA:} reasoning about physical commonsense in natural language.
\newblock In \emph{The Thirty-Fourth {AAAI} Conference on Artificial Intelligence, {AAAI} 2020, The Thirty-Second Innovative Applications of Artificial Intelligence Conference, {IAAI} 2020, The Tenth {AAAI} Symposium on Educational Advances in Artificial Intelligence, {EAAI} 2020, New York, NY, USA, February 7-12, 2020}, pages 7432--7439. {AAAI} Press, 2020.
\newblock \doi{10.1609/AAAI.V34I05.6239}.
\newblock \url{https://doi.org/10.1609/aaai.v34i05.6239}.

\bibitem[Chen et~al.(2021)Chen, Tworek, Jun, Yuan, de~Oliveira~Pinto, Kaplan, Edwards, Burda, Joseph, Brockman, Ray, Puri, Krueger, Petrov, Khlaaf, Sastry, Mishkin, Chan, Gray, Ryder, Pavlov, Power, Kaiser, Bavarian, Winter, Tillet, Such, Cummings, Plappert, Chantzis, Barnes, Herbert{-}Voss, Guss, Nichol, Paino, Tezak, Tang, Babuschkin, Balaji, Jain, Saunders, Hesse, Carr, Leike, Achiam, Misra, Morikawa, Radford, Knight, Brundage, Murati, Mayer, Welinder, McGrew, Amodei, McCandlish, Sutskever, and Zaremba]{humaneval}
Mark Chen, Jerry Tworek, Heewoo Jun, Qiming Yuan, Henrique~Pond{\'{e}} de~Oliveira~Pinto, Jared Kaplan, Harri Edwards, Yuri Burda, Nicholas Joseph, Greg Brockman, Alex Ray, Raul Puri, Gretchen Krueger, Michael Petrov, Heidy Khlaaf, Girish Sastry, Pamela Mishkin, Brooke Chan, Scott Gray, Nick Ryder, Mikhail Pavlov, Alethea Power, Lukasz Kaiser, Mohammad Bavarian, Clemens Winter, Philippe Tillet, Felipe~Petroski Such, Dave Cummings, Matthias Plappert, Fotios Chantzis, Elizabeth Barnes, Ariel Herbert{-}Voss, William~Hebgen Guss, Alex Nichol, Alex Paino, Nikolas Tezak, Jie Tang, Igor Babuschkin, Suchir Balaji, Shantanu Jain, William Saunders, Christopher Hesse, Andrew~N. Carr, Jan Leike, Joshua Achiam, Vedant Misra, Evan Morikawa, Alec Radford, Matthew Knight, Miles Brundage, Mira Murati, Katie Mayer, Peter Welinder, Bob McGrew, Dario Amodei, Sam McCandlish, Ilya Sutskever, and Wojciech Zaremba.
\newblock Evaluating large language models trained on code.
\newblock \emph{CoRR}, abs/2107.03374, 2021.
\newblock \url{https://arxiv.org/abs/2107.03374}.

\bibitem[Cobbe et~al.(2021)Cobbe, Kosaraju, Bavarian, Chen, Jun, Kaiser, Plappert, Tworek, Hilton, Nakano, Hesse, and Schulman]{gsm8k}
Karl Cobbe, Vineet Kosaraju, Mohammad Bavarian, Mark Chen, Heewoo Jun, Lukasz Kaiser, Matthias Plappert, Jerry Tworek, Jacob Hilton, Reiichiro Nakano, Christopher Hesse, and John Schulman.
\newblock Training verifiers to solve math word problems.
\newblock \emph{CoRR}, abs/2110.14168, 2021.
\newblock \url{https://arxiv.org/abs/2110.14168}.

\bibitem[DeepSeek{-}AI et~al.(2024)DeepSeek{-}AI, Liu, Feng, Xue, Wang, Wu, Lu, Zhao, Deng, Zhang, Ruan, Dai, Guo, Yang, Chen, Ji, Li, Lin, Dai, Luo, Hao, Chen, Li, Zhang, Bao, Xu, Wang, Zhang, Ding, Xin, Gao, Li, Qu, Cai, Liang, Guo, Ni, Li, Wang, Chen, Chen, Yuan, Qiu, Li, Song, Dong, Hu, Gao, Guan, Huang, Yu, Wang, Zhang, Xu, Xia, Zhao, Wang, Zhang, Li, Wang, Zhang, Zhang, Tang, Li, Tian, Huang, Wang, Zhang, Wang, Zhu, Chen, Du, Chen, Jin, Ge, Zhang, Pan, Wang, Xu, Zhang, Chen, Li, Lu, Zhou, Chen, Wu, Ye, Ye, Ma, Wang, Zhou, Yu, Zhou, Pan, Wang, Yun, Pei, Sun, Xiao, and Zeng]{deepseekv3}
DeepSeek{-}AI, Aixin Liu, Bei Feng, Bing Xue, Bingxuan Wang, Bochao Wu, Chengda Lu, Chenggang Zhao, Chengqi Deng, Chenyu Zhang, Chong Ruan, Damai Dai, Daya Guo, Dejian Yang, Deli Chen, Dongjie Ji, Erhang Li, Fangyun Lin, Fucong Dai, Fuli Luo, Guangbo Hao, Guanting Chen, Guowei Li, H.~Zhang, Han Bao, Hanwei Xu, Haocheng Wang, Haowei Zhang, Honghui Ding, Huajian Xin, Huazuo Gao, Hui Li, Hui Qu, J.~L. Cai, Jian Liang, Jianzhong Guo, Jiaqi Ni, Jiashi Li, Jiawei Wang, Jin Chen, Jingchang Chen, Jingyang Yuan, Junjie Qiu, Junlong Li, Junxiao Song, Kai Dong, Kai Hu, Kaige Gao, Kang Guan, Kexin Huang, Kuai Yu, Lean Wang, Lecong Zhang, Lei Xu, Leyi Xia, Liang Zhao, Litong Wang, Liyue Zhang, Meng Li, Miaojun Wang, Mingchuan Zhang, Minghua Zhang, Minghui Tang, Mingming Li, Ning Tian, Panpan Huang, Peiyi Wang, Peng Zhang, Qiancheng Wang, Qihao Zhu, Qinyu Chen, Qiushi Du, R.~J. Chen, R.~L. Jin, Ruiqi Ge, Ruisong Zhang, Ruizhe Pan, Runji Wang, Runxin Xu, Ruoyu Zhang, Ruyi Chen, S.~S. Li, Shanghao Lu, Shangyan Zhou,
  Shanhuang Chen, Shaoqing Wu, Shengfeng Ye, Shengfeng Ye, Shirong Ma, Shiyu Wang, Shuang Zhou, Shuiping Yu, Shunfeng Zhou, Shuting Pan, T.~Wang, Tao Yun, Tian Pei, Tianyu Sun, W.~L. Xiao, and Wangding Zeng.
\newblock Deepseek-v3 technical report.
\newblock \emph{CoRR}, abs/2412.19437, 2024.
\newblock \doi{10.48550/ARXIV.2412.19437}.
\newblock \url{https://doi.org/10.48550/arXiv.2412.19437}.

\bibitem[Defazio et~al.(2023)Defazio, Cutkosky, Mehta, and Mishchenko]{defazio2023optimal}
Aaron Defazio, Ashok Cutkosky, Harsh Mehta, and Konstantin Mishchenko.
\newblock Optimal linear decay learning rate schedules and further refinements.
\newblock \emph{arXiv preprint arXiv:2310.07831}, 2023.

\bibitem[Defazio et~al.(2024)Defazio, Yang, Khaled, Mishchenko, Mehta, and Cutkosky]{Aaron2024Road}
Aaron Defazio, Xingyu Yang, Ahmed Khaled, Konstantin Mishchenko, Harsh Mehta, and Ashok Cutkosky.
\newblock The road less scheduled.
\newblock In \emph{NeurIPS}, 2024.

\bibitem[ERNIE-Team(2025)]{ernie2025technicalreport}
Baidu ERNIE-Team.
\newblock Ernie 4.5 technical report, 2025.

\bibitem[Gotmare et~al.(2019)Gotmare, Keskar, Xiong, and Socher]{Akhilesh}
Akhilesh Gotmare, Nitish~Shirish Keskar, Caiming Xiong, and Richard Socher.
\newblock A closer look at deep learning heuristics: Learning rate restarts, warmup and distillation.
\newblock In \emph{7th International Conference on Learning Representations, {ICLR} 2019, New Orleans, LA, USA, May 6-9, 2019}. OpenReview.net, 2019.
\newblock \url{https://openreview.net/forum?id=r14EOsCqKX}.

\bibitem[Grattafiori et~al.(2024)Grattafiori, Dubey, Jauhri, Pandey, Kadian, Al-Dahle, Letman, Mathur, Schelten, Vaughan, et~al.]{grattafiori2024llama}
Aaron Grattafiori, Abhimanyu Dubey, Abhinav Jauhri, Abhinav Pandey, Abhishek Kadian, Ahmad Al-Dahle, Aiesha Letman, Akhil Mathur, Alan Schelten, Alex Vaughan, et~al.
\newblock The llama 3 herd of models.
\newblock \emph{arXiv preprint arXiv:2407.21783}, 2024.

\bibitem[Gu et~al.(2024)Gu, Rozi{\`{e}}re, Leather, Solar{-}Lezama, Synnaeve, and Wang]{cruxeval}
Alex Gu, Baptiste Rozi{\`{e}}re, Hugh~James Leather, Armando Solar{-}Lezama, Gabriel Synnaeve, and Sida Wang.
\newblock Cruxeval: {A} benchmark for code reasoning, understanding and execution.
\newblock In \emph{Forty-first International Conference on Machine Learning, {ICML} 2024, Vienna, Austria, July 21-27, 2024}. OpenReview.net, 2024.
\newblock \url{https://openreview.net/forum?id=Ffpg52swvg}.

\bibitem[H{\"{a}}gele et~al.(2024)H{\"{a}}gele, Bakouch, Kosson, Allal, von Werra, and Jaggi]{1sqrt}
Alexander H{\"{a}}gele, Elie Bakouch, Atli Kosson, Loubna~Ben Allal, Leandro von Werra, and Martin Jaggi.
\newblock Scaling laws and compute-optimal training beyond fixed training durations.
\newblock In \emph{NeurIPS}, 2024.

\bibitem[Hendrycks et~al.(2021{\natexlab{a}})Hendrycks, Burns, Basart, Zou, Mazeika, Song, and Steinhardt]{mmlu}
Dan Hendrycks, Collin Burns, Steven Basart, Andy Zou, Mantas Mazeika, Dawn Song, and Jacob Steinhardt.
\newblock Measuring massive multitask language understanding.
\newblock In \emph{9th International Conference on Learning Representations, {ICLR} 2021, Virtual Event, Austria, May 3-7, 2021}. OpenReview.net, 2021{\natexlab{a}}.
\newblock \url{https://openreview.net/forum?id=d7KBjmI3GmQ}.

\bibitem[Hendrycks et~al.(2021{\natexlab{b}})Hendrycks, Burns, Kadavath, Arora, Basart, Tang, Song, and Steinhardt]{math}
Dan Hendrycks, Collin Burns, Saurav Kadavath, Akul Arora, Steven Basart, Eric Tang, Dawn Song, and Jacob Steinhardt.
\newblock Measuring mathematical problem solving with the {MATH} dataset.
\newblock In Joaquin Vanschoren and Sai{-}Kit Yeung, editors, \emph{Proceedings of the Neural Information Processing Systems Track on Datasets and Benchmarks 1, NeurIPS Datasets and Benchmarks 2021, December 2021, virtual}, 2021{\natexlab{b}}.
\newblock \url{https://datasets-benchmarks-proceedings.neurips.cc/paper/2021/hash/be83ab3ecd0db773eb2dc1b0a17836a1-Abstract-round2.html}.

\bibitem[Hoffmann et~al.(2022)Hoffmann, Borgeaud, Mensch, Buchatskaya, Cai, Rutherford, Casas, Hendricks, Welbl, Clark, et~al.]{hoffmann2022training}
Jordan Hoffmann, Sebastian Borgeaud, Arthur Mensch, Elena Buchatskaya, Trevor Cai, Eliza Rutherford, Diego de~Las Casas, Lisa~Anne Hendricks, Johannes Welbl, Aidan Clark, et~al.
\newblock Training compute-optimal large language models.
\newblock \emph{arXiv preprint arXiv:2203.15556}, 2022.

\bibitem[Hong et~al.(2025)Hong, Yan, Cai, Jiang, Hu, and Xie]{worldsense}
Jack Hong, Shilin Yan, Jiayin Cai, Xiaolong Jiang, Yao Hu, and Weidi Xie.
\newblock Worldsense: Evaluating real-world omnimodal understanding for multimodal llms.
\newblock \emph{CoRR}, abs/2502.04326, 2025.
\newblock \doi{10.48550/ARXIV.2502.04326}.
\newblock \url{https://doi.org/10.48550/arXiv.2502.04326}.

\bibitem[Hu et~al.(2024)Hu, Tu, Han, He, Cui, Long, Zheng, Fang, Huang, Zhao, Zhang, Thai, Zhang, Wang, Yao, Zhao, Zhou, Cai, Zhai, Ding, Jia, Zeng, Li, Liu, and Sun]{minicpm}
Shengding Hu, Yuge Tu, Xu~Han, Chaoqun He, Ganqu Cui, Xiang Long, Zhi Zheng, Yewei Fang, Yuxiang Huang, Weilin Zhao, Xinrong Zhang, Zhen~Leng Thai, Kai Zhang, Chongyi Wang, Yuan Yao, Chenyang Zhao, Jie Zhou, Jie Cai, Zhongwu Zhai, Ning Ding, Chao Jia, Guoyang Zeng, Dahai Li, Zhiyuan Liu, and Maosong Sun.
\newblock Minicpm: Unveiling the potential of small language models with scalable training strategies.
\newblock \emph{CoRR}, abs/2404.06395, 2024.
\newblock \doi{10.48550/ARXIV.2404.06395}.
\newblock \url{https://doi.org/10.48550/arXiv.2404.06395}.

\bibitem[Huang et~al.(2023)Huang, Bai, Zhu, Zhang, Zhang, Su, Liu, Lv, Zhang, Lei, Fu, Sun, and He]{ceval}
Yuzhen Huang, Yuzhuo Bai, Zhihao Zhu, Junlei Zhang, Jinghan Zhang, Tangjun Su, Junteng Liu, Chuancheng Lv, Yikai Zhang, Jiayi Lei, Yao Fu, Maosong Sun, and Junxian He.
\newblock C-eval: {A} multi-level multi-discipline chinese evaluation suite for foundation models.
\newblock In Alice Oh, Tristan Naumann, Amir Globerson, Kate Saenko, Moritz Hardt, and Sergey Levine, editors, \emph{Advances in Neural Information Processing Systems 36: Annual Conference on Neural Information Processing Systems 2023, NeurIPS 2023, New Orleans, LA, USA, December 10 - 16, 2023}, 2023.
\newblock \url{http://papers.nips.cc/paper\_files/paper/2023/hash/c6ec1844bec96d6d32ae95ae694e23d8-Abstract-Datasets\_and\_Benchmarks.html}.

\bibitem[Ibrahim et~al.(2024)Ibrahim, Th{\'{e}}rien, Gupta, Richter, Anthony, Belilovsky, Lesort, and Rish]{Simple}
Adam Ibrahim, Benjamin Th{\'{e}}rien, Kshitij Gupta, Mats~L. Richter, Quentin~Gregory Anthony, Eugene Belilovsky, Timoth{\'{e}}e Lesort, and Irina Rish.
\newblock Simple and scalable strategies to continually pre-train large language models.
\newblock \emph{Trans. Mach. Learn. Res.}, 2024, 2024.
\newblock \url{https://openreview.net/forum?id=DimPeeCxKO}.

\bibitem[Izmailov et~al.(2018)Izmailov, Podoprikhin, Garipov, Vetrov, and Wilson]{Averaging}
Pavel Izmailov, Dmitrii Podoprikhin, Timur Garipov, Dmitry~P. Vetrov, and Andrew~Gordon Wilson.
\newblock Averaging weights leads to wider optima and better generalization.
\newblock In Amir Globerson and Ricardo Silva, editors, \emph{Proceedings of the Thirty-Fourth Conference on Uncertainty in Artificial Intelligence, {UAI} 2018, Monterey, California, USA, August 6-10, 2018}, pages 876--885. {AUAI} Press, 2018.
\newblock \url{http://auai.org/uai2018/proceedings/papers/313.pdf}.

\bibitem[Jain et~al.(2025)Jain, Han, Gu, Li, Yan, Zhang, Wang, Solar{-}Lezama, Sen, and Stoica]{livecodebench}
Naman Jain, King Han, Alex Gu, Wen{-}Ding Li, Fanjia Yan, Tianjun Zhang, Sida Wang, Armando Solar{-}Lezama, Koushik Sen, and Ion Stoica.
\newblock Livecodebench: Holistic and contamination free evaluation of large language models for code.
\newblock In \emph{The Thirteenth International Conference on Learning Representations, {ICLR} 2025, Singapore, April 24-28, 2025}. OpenReview.net, 2025.
\newblock \url{https://openreview.net/forum?id=chfJJYC3iL}.

\bibitem[Jin et~al.(2023)Jin, Wei, Wang, Zhang, and Wu]{Rethinking}
Hongpeng Jin, Wenqi Wei, Xuyu Wang, Wenbin Zhang, and Yanzhao Wu.
\newblock Rethinking learning rate tuning in the era of large language models.
\newblock In \emph{5th {IEEE} International Conference on Cognitive Machine Intelligence, CogMI 2023, Atlanta, GA, USA, November 1-4, 2023}, pages 112--121. {IEEE}, 2023.
\newblock \doi{10.1109/COGMI58952.2023.00025}.
\newblock \url{https://doi.org/10.1109/CogMI58952.2023.00025}.

\bibitem[Joshi et~al.(2017)Joshi, Choi, Weld, and Zettlemoyer]{trivalqa}
Mandar Joshi, Eunsol Choi, Daniel~S. Weld, and Luke Zettlemoyer.
\newblock Triviaqa: {A} large scale distantly supervised challenge dataset for reading comprehension.
\newblock In Regina Barzilay and Min{-}Yen Kan, editors, \emph{Proceedings of the 55th Annual Meeting of the Association for Computational Linguistics, {ACL} 2017, Vancouver, Canada, July 30 - August 4, Volume 1: Long Papers}, pages 1601--1611. Association for Computational Linguistics, 2017.
\newblock \doi{10.18653/V1/P17-1147}.
\newblock \url{https://doi.org/10.18653/v1/P17-1147}.

\bibitem[Kaddour(2022)]{lawa}
Jean Kaddour.
\newblock Stop wasting my time! saving days of imagenet and {BERT} training with latest weight averaging.
\newblock \emph{CoRR}, abs/2209.14981, 2022.
\newblock \doi{10.48550/ARXIV.2209.14981}.
\newblock \url{https://doi.org/10.48550/arXiv.2209.14981}.

\bibitem[Kaplan et~al.(2020)Kaplan, McCandlish, Henighan, Brown, Chess, Child, Gray, Radford, Wu, and Amodei]{kaplan2020scaling}
Jared Kaplan, Sam McCandlish, Tom Henighan, Tom~B Brown, Benjamin Chess, Rewon Child, Scott Gray, Alec Radford, Jeffrey Wu, and Dario Amodei.
\newblock Scaling laws for neural language models.
\newblock \emph{arXiv preprint arXiv:2001.08361}, 2020.

\bibitem[Kingma and Ba(2015)]{adam}
Diederik~P. Kingma and Jimmy Ba.
\newblock Adam: {A} method for stochastic optimization.
\newblock In Yoshua Bengio and Yann LeCun, editors, \emph{3rd International Conference on Learning Representations, {ICLR} 2015, San Diego, CA, USA, May 7-9, 2015, Conference Track Proceedings}, 2015.
\newblock \url{http://arxiv.org/abs/1412.6980}.

\bibitem[Kwiatkowski et~al.(2019)Kwiatkowski, Palomaki, Redfield, Collins, Parikh, Alberti, Epstein, Polosukhin, Devlin, Lee, Toutanova, Jones, Kelcey, Chang, Dai, Uszkoreit, Le, and Petrov]{nq}
Tom Kwiatkowski, Jennimaria Palomaki, Olivia Redfield, Michael Collins, Ankur~P. Parikh, Chris Alberti, Danielle Epstein, Illia Polosukhin, Jacob Devlin, Kenton Lee, Kristina Toutanova, Llion Jones, Matthew Kelcey, Ming{-}Wei Chang, Andrew~M. Dai, Jakob Uszkoreit, Quoc Le, and Slav Petrov.
\newblock Natural questions: a benchmark for question answering research.
\newblock \emph{Trans. Assoc. Comput. Linguistics}, 7:\penalty0 452--466, 2019.
\newblock \doi{10.1162/TACL\_A\_00276}.
\newblock \url{https://doi.org/10.1162/tacl\_a\_00276}.

\bibitem[Lai et~al.(2017)Lai, Xie, Liu, Yang, and Hovy]{race}
Guokun Lai, Qizhe Xie, Hanxiao Liu, Yiming Yang, and Eduard~H. Hovy.
\newblock {RACE:} large-scale reading comprehension dataset from examinations.
\newblock In Martha Palmer, Rebecca Hwa, and Sebastian Riedel, editors, \emph{Proceedings of the 2017 Conference on Empirical Methods in Natural Language Processing, {EMNLP} 2017, Copenhagen, Denmark, September 9-11, 2017}, pages 785--794. Association for Computational Linguistics, 2017.
\newblock \doi{10.18653/V1/D17-1082}.
\newblock \url{https://doi.org/10.18653/v1/d17-1082}.

\bibitem[Li et~al.(2024{\natexlab{a}})Li, Zhang, Koto, Yang, Zhao, Gong, Duan, and Baldwin]{cmmlu}
Haonan Li, Yixuan Zhang, Fajri Koto, Yifei Yang, Hai Zhao, Yeyun Gong, Nan Duan, and Timothy Baldwin.
\newblock {CMMLU:} measuring massive multitask language understanding in chinese.
\newblock In Lun{-}Wei Ku, Andre Martins, and Vivek Srikumar, editors, \emph{Findings of the Association for Computational Linguistics, {ACL} 2024, Bangkok, Thailand and virtual meeting, August 11-16, 2024}, pages 11260--11285. Association for Computational Linguistics, 2024{\natexlab{a}}.
\newblock \doi{10.18653/V1/2024.FINDINGS-ACL.671}.
\newblock \url{https://doi.org/10.18653/v1/2024.findings-acl.671}.

\bibitem[Li et~al.(2024{\natexlab{b}})Li, Cui, Zhao, Kong, and Bi]{gsm-plus}
Qintong Li, Leyang Cui, Xueliang Zhao, Lingpeng Kong, and Wei Bi.
\newblock Gsm-plus: {A} comprehensive benchmark for evaluating the robustness of llms as mathematical problem solvers.
\newblock In Lun{-}Wei Ku, Andre Martins, and Vivek Srikumar, editors, \emph{Proceedings of the 62nd Annual Meeting of the Association for Computational Linguistics (Volume 1: Long Papers), {ACL} 2024, Bangkok, Thailand, August 11-16, 2024}, pages 2961--2984. Association for Computational Linguistics, 2024{\natexlab{b}}.
\newblock \doi{10.18653/V1/2024.ACL-LONG.163}.
\newblock \url{https://doi.org/10.18653/v1/2024.acl-long.163}.

\bibitem[Li et~al.(2023)Li, Huang, Tao, Wu, and Huang]{Trainable}
Tao Li, Zhehao Huang, Qinghua Tao, Yingwen Wu, and Xiaolin Huang.
\newblock Trainable weight averaging: Efficient training by optimizing historical solutions.
\newblock In \emph{The Eleventh International Conference on Learning Representations, {ICLR} 2023, Kigali, Rwanda, May 1-5, 2023}. OpenReview.net, 2023.
\newblock \url{https://openreview.net/forum?id=8wbnpOJY-f}.

\bibitem[Li et~al.(2025)Li, Ma, Yan, Zhang, Liu, Lu, Xu, Chen, Wang, Zhan, Ma, Lai, Liu, Luo, Bin, Ren, Han, Hao, Yi, Liu, Ma, Jia, Zhou, Qiao, Xiang, and Wu]{pma}
Yunshui Li, Yiyuan Ma, Shen Yan, Chaoyi Zhang, Jing Liu, Jianqiao Lu, Ziwen Xu, Mengzhao Chen, Minrui Wang, Shiyi Zhan, Jin Ma, Xunhao Lai, Deyi Liu, Yao Luo, Xingyan Bin, Hongbin Ren, Mingji Han, Wenhao Hao, Bairen Yi, LingJun Liu, Bole Ma, Xiaoying Jia, Xun Zhou, Siyuan Qiao, Liang Xiang, and Yonghui Wu.
\newblock Model merging in pre-training of large language models.
\newblock \emph{CoRR}, abs/2505.12082, 2025.
\newblock \doi{10.48550/ARXIV.2505.12082}.
\newblock \url{https://doi.org/10.48550/arXiv.2505.12082}.

\bibitem[Ling-Team et~al.(2025)Ling-Team, Zeng, Huang, Zhang, Tian, Chen, Jin, Yu, Zhu, Yuan, et~al.]{team2025every}
Ling-Team, Binwei Zeng, Chao Huang, Chao Zhang, Changxin Tian, Cong Chen, Dingnan Jin, Feng Yu, Feng Zhu, Feng Yuan, et~al.
\newblock Every flop counts: Scaling a 300b mixture-of-experts ling llm without premium gpus.
\newblock \emph{arXiv preprint arXiv:2503.05139}, 2025.

\bibitem[Liu et~al.(2024{\natexlab{a}})Liu, Wang, Wang, Chen, Li, Tu, Chu, Li, and Sui]{Checkpoint}
Deyuan Liu, Zecheng Wang, Bingning Wang, Weipeng Chen, Chunshan Li, Zhiying Tu, Dianhui Chu, Bo~Li, and Dianbo Sui.
\newblock Checkpoint merging via bayesian optimization in {LLM} pretraining.
\newblock \emph{CoRR}, abs/2403.19390, 2024{\natexlab{a}}.
\newblock \doi{10.48550/ARXIV.2403.19390}.
\newblock \url{https://doi.org/10.48550/arXiv.2403.19390}.

\bibitem[Liu et~al.(2024{\natexlab{b}})Liu, Zheng, Qiao, Duan, Fei, Zhou, Zhang, Zhang, Lin, and Chen]{mathbench}
Hongwei Liu, Zilong Zheng, Yuxuan Qiao, Haodong Duan, Zhiwei Fei, Fengzhe Zhou, Wenwei Zhang, Songyang Zhang, Dahua Lin, and Kai Chen.
\newblock Mathbench: Evaluating the theory and application proficiency of llms with a hierarchical mathematics benchmark.
\newblock In Lun{-}Wei Ku, Andre Martins, and Vivek Srikumar, editors, \emph{Findings of the Association for Computational Linguistics, {ACL} 2024, Bangkok, Thailand and virtual meeting, August 11-16, 2024}, pages 6884--6915. Association for Computational Linguistics, 2024{\natexlab{b}}.
\newblock \doi{10.18653/V1/2024.FINDINGS-ACL.411}.
\newblock \url{https://doi.org/10.18653/v1/2024.findings-acl.411}.

\bibitem[Liu et~al.(2023)Liu, Xia, Wang, and Zhang]{mbpp+}
Jiawei Liu, Chunqiu~Steven Xia, Yuyao Wang, and Lingming Zhang.
\newblock Is your code generated by chatgpt really correct? rigorous evaluation of large language models for code generation.
\newblock In Alice Oh, Tristan Naumann, Amir Globerson, Kate Saenko, Moritz Hardt, and Sergey Levine, editors, \emph{Advances in Neural Information Processing Systems 36: Annual Conference on Neural Information Processing Systems 2023, NeurIPS 2023, New Orleans, LA, USA, December 10 - 16, 2023}, 2023.
\newblock \url{http://papers.nips.cc/paper\_files/paper/2023/hash/43e9d647ccd3e4b7b5baab53f0368686-Abstract-Conference.html}.

\bibitem[Liu et~al.(2025)Liu, Su, Yao, Jiang, Lai, Du, Qin, Xu, Lu, Yan, Chen, Zheng, Liu, Liu, Yin, He, Zhu, Wang, Wang, Dong, Zhang, Kang, Zhang, Xu, Zhang, Wu, Zhou, and Yang]{liu2025muonscalablellmtraining}
Jingyuan Liu, Jianlin Su, Xingcheng Yao, Zhejun Jiang, Guokun Lai, Yulun Du, Yidao Qin, Weixin Xu, Enzhe Lu, Junjie Yan, Yanru Chen, Huabin Zheng, Yibo Liu, Shaowei Liu, Bohong Yin, Weiran He, Han Zhu, Yuzhi Wang, Jianzhou Wang, Mengnan Dong, Zheng Zhang, Yongsheng Kang, Hao Zhang, Xinran Xu, Yutao Zhang, Yuxin Wu, Xinyu Zhou, and Zhilin Yang.
\newblock Muon is scalable for llm training, 2025.
\newblock \url{https://arxiv.org/abs/2502.16982}.

\bibitem[Loshchilov and Hutter(2019)]{adamw}
Ilya Loshchilov and Frank Hutter.
\newblock Decoupled weight decay regularization.
\newblock In \emph{7th International Conference on Learning Representations, {ICLR} 2019, New Orleans, LA, USA, May 6-9, 2019}. OpenReview.net, 2019.
\newblock \url{https://openreview.net/forum?id=Bkg6RiCqY7}.

\bibitem[Ma et~al.(2025)Ma, Du, Wang, Zhang, Wen, Qu, Yang, Liu, Liu, Yue, Huang, and Zhang]{korbench}
Kaijing Ma, Xeron Du, Yunran Wang, Haoran Zhang, Zhoufutu Wen, Xingwei Qu, Jian Yang, Jiaheng Liu, Minghao Liu, Xiang Yue, Wenhao Huang, and Ge~Zhang.
\newblock Kor-bench: Benchmarking language models on knowledge-orthogonal reasoning tasks.
\newblock In \emph{The Thirteenth International Conference on Learning Representations, {ICLR} 2025, Singapore, April 24-28, 2025}. OpenReview.net, 2025.
\newblock \url{https://openreview.net/forum?id=SVRRQ8goQo}.

\bibitem[Mihaylov et~al.(2018)Mihaylov, Clark, Khot, and Sabharwal]{openbookqa}
Todor Mihaylov, Peter Clark, Tushar Khot, and Ashish Sabharwal.
\newblock Can a suit of armor conduct electricity? {A} new dataset for open book question answering.
\newblock In Ellen Riloff, David Chiang, Julia Hockenmaier, and Jun'ichi Tsujii, editors, \emph{Proceedings of the 2018 Conference on Empirical Methods in Natural Language Processing, Brussels, Belgium, October 31 - November 4, 2018}, pages 2381--2391. Association for Computational Linguistics, 2018.
\newblock \doi{10.18653/V1/D18-1260}.
\newblock \url{https://doi.org/10.18653/v1/d18-1260}.

\bibitem[Peng et~al.(2024)Peng, Chai, and Li]{humaneval_cn}
Qiwei Peng, Yekun Chai, and Xuhong Li.
\newblock Humaneval-xl: {A} multilingual code generation benchmark for cross-lingual natural language generalization.
\newblock In Nicoletta Calzolari, Min{-}Yen Kan, V{\'{e}}ronique Hoste, Alessandro Lenci, Sakriani Sakti, and Nianwen Xue, editors, \emph{Proceedings of the 2024 Joint International Conference on Computational Linguistics, Language Resources and Evaluation, {LREC/COLING} 2024, 20-25 May, 2024, Torino, Italy}, pages 8383--8394. {ELRA} and {ICCL}, 2024.
\newblock \url{https://aclanthology.org/2024.lrec-main.735}.

\bibitem[Polyak and Juditsky(1992)]{polyak1992acceleration}
Boris~T Polyak and Anatoli~B Juditsky.
\newblock Acceleration of stochastic approximation by averaging.
\newblock \emph{SIAM journal on control and optimization}, 30\penalty0 (4):\penalty0 838--855, 1992.

\bibitem[Rajpurkar et~al.(2018)Rajpurkar, Jia, and Liang]{squad2}
Pranav Rajpurkar, Robin Jia, and Percy Liang.
\newblock Know what you don't know: Unanswerable questions for squad.
\newblock In Iryna Gurevych and Yusuke Miyao, editors, \emph{Proceedings of the 56th Annual Meeting of the Association for Computational Linguistics, {ACL} 2018, Melbourne, Australia, July 15-20, 2018, Volume 2: Short Papers}, pages 784--789. Association for Computational Linguistics, 2018.
\newblock \doi{10.18653/V1/P18-2124}.
\newblock \url{https://aclanthology.org/P18-2124/}.

\bibitem[Ram{\'{e}} et~al.(2024)Ram{\'{e}}, Ferret, Vieillard, Dadashi, Hussenot, Cedoz, Sessa, Girgin, Douillard, and Bachem]{WARP}
Alexandre Ram{\'{e}}, Johan Ferret, Nino Vieillard, Robert Dadashi, L{\'{e}}onard Hussenot, Pierre{-}Louis Cedoz, Pier~Giuseppe Sessa, Sertan Girgin, Arthur Douillard, and Olivier Bachem.
\newblock {WARP:} on the benefits of weight averaged rewarded policies.
\newblock \emph{CoRR}, abs/2406.16768, 2024.
\newblock \doi{10.48550/ARXIV.2406.16768}.
\newblock \url{https://doi.org/10.48550/arXiv.2406.16768}.

\bibitem[Rein et~al.(2023)Rein, Hou, Stickland, Petty, Pang, Dirani, Michael, and Bowman]{gpqa}
David Rein, Betty~Li Hou, Asa~Cooper Stickland, Jackson Petty, Richard~Yuanzhe Pang, Julien Dirani, Julian Michael, and Samuel~R. Bowman.
\newblock {GPQA:} {A} graduate-level google-proof q{\&}a benchmark.
\newblock \emph{CoRR}, abs/2311.12022, 2023.
\newblock \doi{10.48550/ARXIV.2311.12022}.
\newblock \url{https://doi.org/10.48550/arXiv.2311.12022}.

\bibitem[Roy and Vetterli(2007)]{roy2007effective}
Olivier Roy and Martin Vetterli.
\newblock The effective rank: A measure of effective dimensionality.
\newblock In \emph{2007 15th European signal processing conference}, pages 606--610. IEEE, 2007.

\bibitem[Sakaguchi et~al.(2021)Sakaguchi, Bras, Bhagavatula, and Choi]{winogrande}
Keisuke Sakaguchi, Ronan~Le Bras, Chandra Bhagavatula, and Yejin Choi.
\newblock Winogrande: an adversarial winograd schema challenge at scale.
\newblock \emph{Commun. {ACM}}, 64\penalty0 (9):\penalty0 99--106, 2021.
\newblock \doi{10.1145/3474381}.
\newblock \url{https://doi.org/10.1145/3474381}.

\bibitem[Sandler et~al.(2023)Sandler, Zhmoginov, Vladymyrov, and Miller]{Sandler2023training}
Mark Sandler, Andrey Zhmoginov, Max Vladymyrov, and Nolan Miller.
\newblock Training trajectories, mini-batch losses and the curious role of the learning rate.
\newblock \emph{CoRR}, abs/2301.02312, 2023.

\bibitem[Sanyal et~al.(2023)Sanyal, Neerkaje, Kaddour, Kumar, and Sanghavi]{sanyal2023early}
Sunny Sanyal, Atula Neerkaje, Jean Kaddour, Abhishek Kumar, and Sujay Sanghavi.
\newblock Early weight averaging meets high learning rates for llm pre-training.
\newblock \emph{arXiv preprint arXiv:2306.03241}, 2023.

\bibitem[Shi et~al.(2023)Shi, Suzgun, Freitag, Wang, Srivats, Vosoughi, Chung, Tay, Ruder, Zhou, Das, and Wei]{mgsm}
Freda Shi, Mirac Suzgun, Markus Freitag, Xuezhi Wang, Suraj Srivats, Soroush Vosoughi, Hyung~Won Chung, Yi~Tay, Sebastian Ruder, Denny Zhou, Dipanjan Das, and Jason Wei.
\newblock Language models are multilingual chain-of-thought reasoners.
\newblock In \emph{The Eleventh International Conference on Learning Representations, {ICLR} 2023, Kigali, Rwanda, May 1-5, 2023}. OpenReview.net, 2023.
\newblock \url{https://openreview.net/forum?id=fR3wGCk-IXp}.

\bibitem[Song et~al.(2025)Song, Baek, Ahn, and Yun]{song2025through}
Minhak Song, Beomhan Baek, Kwangjun Ahn, and Chulhee Yun.
\newblock Through the river: Understanding the benefit of schedule-free methods for language model training.
\newblock In \emph{High-dimensional Learning Dynamics 2025}, 2025.

\bibitem[Su et~al.(2024)Su, Ahmed, Lu, Pan, Bo, and Liu]{rope}
Jianlin Su, Murtadha H.~M. Ahmed, Yu~Lu, Shengfeng Pan, Wen Bo, and Yunfeng Liu.
\newblock Roformer: Enhanced transformer with rotary position embedding.
\newblock \emph{Neurocomputing}, 568:\penalty0 127063, 2024.
\newblock \doi{10.1016/J.NEUCOM.2023.127063}.
\newblock \url{https://doi.org/10.1016/j.neucom.2023.127063}.

\bibitem[Suzgun et~al.(2023)Suzgun, Scales, Sch{\"{a}}rli, Gehrmann, Tay, Chung, Chowdhery, Le, Chi, Zhou, and Wei]{bbh}
Mirac Suzgun, Nathan Scales, Nathanael Sch{\"{a}}rli, Sebastian Gehrmann, Yi~Tay, Hyung~Won Chung, Aakanksha Chowdhery, Quoc~V. Le, Ed~H. Chi, Denny Zhou, and Jason Wei.
\newblock Challenging big-bench tasks and whether chain-of-thought can solve them.
\newblock In Anna Rogers, Jordan~L. Boyd{-}Graber, and Naoaki Okazaki, editors, \emph{Findings of the Association for Computational Linguistics: {ACL} 2023, Toronto, Canada, July 9-14, 2023}, pages 13003--13051. Association for Computational Linguistics, 2023.
\newblock \doi{10.18653/V1/2023.FINDINGS-ACL.824}.
\newblock \url{https://doi.org/10.18653/v1/2023.findings-acl.824}.

\bibitem[Tang et~al.(2024)Tang, Zhang, Wang, and Wei]{college-math}
Zhengyang Tang, Xingxing Zhang, Benyou Wang, and Furu Wei.
\newblock Mathscale: Scaling instruction tuning for mathematical reasoning.
\newblock In \emph{Forty-first International Conference on Machine Learning, {ICML} 2024, Vienna, Austria, July 21-27, 2024}. OpenReview.net, 2024.
\newblock \url{https://openreview.net/forum?id=Kjww7ZN47M}.

\bibitem[Tao et~al.(2024)Tao, Ventresque, Nallur, and Saber]{mbpp}
Ning Tao, Anthony Ventresque, Vivek Nallur, and Takfarinas Saber.
\newblock Enhancing program synthesis with large language models using many-objective grammar-guided genetic programming.
\newblock \emph{Algorithms}, 17\penalty0 (7):\penalty0 287, 2024.
\newblock \doi{10.3390/A17070287}.
\newblock \url{https://doi.org/10.3390/a17070287}.

\bibitem[Team et~al.(2025)Team, Du, Yao, Ma, Wang, Zheng, Zhu, Liu, Liang, Jin, Wei, Zheng, Deng, Jia, Jiang, Liao, Li, Li, Li, Li, Li, Ma, Ni, Que, Wang, Wen, Wu, Xing, Xu, Yang, Wang, Zhou, Bai, Bu, Cai, Chen, Chen, Cheng, Cheng, Ding, Huang, Huang, Li, Li, Li, Liang, Lin, Lin, Ma, Pang, Peng, Peng, Qi, Qiu, Qu, Quan, Tan, Wang, Wang, Wang, Wang, Wang, Xu, Yang, Yuan, Yue, Zhan, Zhang, Zhang, Zhang, Zhang, Zhang, Zhao, Zheng, Zhong, Gao, Li, Liu, Liu, Liu, Ni, Peng, Qin, Su, Wang, Wang, Yang, Yang, Cao, Yue, Zhang, Zhou, Liu, Lin, Huang, and Zhang]{supergpqa}
M.{-}A{-}P. Team, Xinrun Du, Yifan Yao, Kaijing Ma, Bingli Wang, Tianyu Zheng, Kang Zhu, Minghao Liu, Yiming Liang, Xiaolong Jin, Zhenlin Wei, Chujie Zheng, Kaixin Deng, Shian Jia, Sichao Jiang, Yiyan Liao, Rui Li, Qinrui Li, Sirun Li, Yizhi Li, Yunwen Li, Dehua Ma, Yuansheng Ni, Haoran Que, Qiyao Wang, Zhoufutu Wen, Siwei Wu, Tianshun Xing, Ming Xu, Zhenzhu Yang, Zekun~Moore Wang, Jun Zhou, Yuelin Bai, Xingyuan Bu, Chenglin Cai, Liang Chen, Yifan Chen, Chengtuo Cheng, Tianhao Cheng, Keyi Ding, Siming Huang, Yun Huang, Yaoru Li, Yizhe Li, Zhaoqun Li, Tianhao Liang, Chengdong Lin, Hongquan Lin, Yinghao Ma, Tianyang Pang, Zhongyuan Peng, Zifan Peng, Qige Qi, Shi Qiu, Xingwei Qu, Shanghaoran Quan, Yizhou Tan, Zili Wang, Chenqing Wang, Hao Wang, Yiya Wang, Yubo Wang, Jiajun Xu, Kexin Yang, Ruibin Yuan, Yuanhao Yue, Tianyang Zhan, Chun Zhang, Jinyang Zhang, Xiyue Zhang, Xingjian Zhang, Yue Zhang, Yongchi Zhao, Xiangyu Zheng, Chenghua Zhong, Yang Gao, Zhoujun Li, Dayiheng Liu, Qian Liu, Tianyu Liu, Shiwen Ni,
  Junran Peng, Yujia Qin, Wenbo Su, Guoyin Wang, Shi Wang, Jian Yang, Min Yang, Meng Cao, Xiang Yue, Zhaoxiang Zhang, Wangchunshu Zhou, Jiaheng Liu, Qunshu Lin, Wenhao Huang, and Ge~Zhang.
\newblock Supergpqa: Scaling {LLM} evaluation across 285 graduate disciplines.
\newblock \emph{CoRR}, abs/2502.14739, 2025.
\newblock \doi{10.48550/ARXIV.2502.14739}.
\newblock \url{https://doi.org/10.48550/arXiv.2502.14739}.

\bibitem[van~der Maaten and Hinton(2008)]{vanDerMaaten2008}
Laurens van~der Maaten and Geoffrey Hinton.
\newblock Visualizing data using {t-SNE}.
\newblock \emph{Journal of Machine Learning Research}, 9:\penalty0 2579--2605, 2008.
\newblock \url{http://www.jmlr.org/papers/v9/vandermaaten08a.html}.

\bibitem[Wang et~al.(2024)Wang, Ma, Zhang, Ni, Chandra, Guo, Ren, Arulraj, He, Jiang, Li, Ku, Wang, Zhuang, Fan, Yue, and Chen]{mmlu-pro}
Yubo Wang, Xueguang Ma, Ge~Zhang, Yuansheng Ni, Abhranil Chandra, Shiguang Guo, Weiming Ren, Aaran Arulraj, Xuan He, Ziyan Jiang, Tianle Li, Max Ku, Kai Wang, Alex Zhuang, Rongqi Fan, Xiang Yue, and Wenhu Chen.
\newblock Mmlu-pro: {A} more robust and challenging multi-task language understanding benchmark.
\newblock In Amir Globersons, Lester Mackey, Danielle Belgrave, Angela Fan, Ulrich Paquet, Jakub~M. Tomczak, and Cheng Zhang, editors, \emph{Advances in Neural Information Processing Systems 38: Annual Conference on Neural Information Processing Systems 2024, NeurIPS 2024, Vancouver, BC, Canada, December 10 - 15, 2024}, 2024.
\newblock \url{http://papers.nips.cc/paper\_files/paper/2024/hash/ad236edc564f3e3156e1b2feafb99a24-Abstract-Datasets\_and\_Benchmarks\_Track.html}.

\bibitem[Wei et~al.(2023)Wei, Luan, Liu, Dong, and Wang]{cmath}
Tianwen Wei, Jian Luan, Wei Liu, Shuang Dong, and Bin Wang.
\newblock {CMATH:} can your language model pass chinese elementary school math test?
\newblock \emph{CoRR}, abs/2306.16636, 2023.
\newblock \doi{10.48550/ARXIV.2306.16636}.
\newblock \url{https://doi.org/10.48550/arXiv.2306.16636}.

\bibitem[Wen et~al.(2024)Wen, Li, Wang, Hall, Liang, and Ma]{wen2024understanding}
Kaiyue Wen, Zhiyuan Li, Jason Wang, David Hall, Percy Liang, and Tengyu Ma.
\newblock Understanding warmup-stable-decay learning rates: A river valley loss landscape perspective.
\newblock \emph{arXiv preprint arXiv:2410.05192}, 2024.

\bibitem[Wortsman et~al.(2022)Wortsman, Ilharco, Gadre, Roelofs, Lopes, Morcos, Namkoong, Farhadi, Carmon, Kornblith, and Schmidt]{Modelsoups}
Mitchell Wortsman, Gabriel Ilharco, Samir~Yitzhak Gadre, Rebecca Roelofs, Raphael~Gontijo Lopes, Ari~S. Morcos, Hongseok Namkoong, Ali Farhadi, Yair Carmon, Simon Kornblith, and Ludwig Schmidt.
\newblock Model soups: averaging weights of multiple fine-tuned models improves accuracy without increasing inference time.
\newblock In Kamalika Chaudhuri, Stefanie Jegelka, Le~Song, Csaba Szepesv{\'{a}}ri, Gang Niu, and Sivan Sabato, editors, \emph{International Conference on Machine Learning, {ICML} 2022, 17-23 July 2022, Baltimore, Maryland, {USA}}, volume 162 of \emph{Proceedings of Machine Learning Research}, pages 23965--23998. {PMLR}, 2022.
\newblock \url{https://proceedings.mlr.press/v162/wortsman22a.html}.

\bibitem[Zellers et~al.(2019)Zellers, Holtzman, Bisk, Farhadi, and Choi]{hellaswag}
Rowan Zellers, Ari Holtzman, Yonatan Bisk, Ali Farhadi, and Yejin Choi.
\newblock Hellaswag: Can a machine really finish your sentence?
\newblock In Anna Korhonen, David~R. Traum, and Llu{\'{\i}}s M{\`{a}}rquez, editors, \emph{Proceedings of the 57th Conference of the Association for Computational Linguistics, {ACL} 2019, Florence, Italy, July 28- August 2, 2019, Volume 1: Long Papers}, pages 4791--4800. Association for Computational Linguistics, 2019.
\newblock \doi{10.18653/V1/P19-1472}.
\newblock \url{https://doi.org/10.18653/v1/p19-1472}.

\bibitem[Zhang et~al.(2025)Zhang, Morwani, Vyas, Wu, Zou, Ghai, Foster, and Kakade]{Hanlin2025cbs}
Hanlin Zhang, Depen Morwani, Nikhil Vyas, Jingfeng Wu, Difan Zou, Udaya Ghai, Dean~P. Foster, and Sham~M. Kakade.
\newblock How does critical batch size scale in pre-training?
\newblock In \emph{{ICLR}}. OpenReview.net, 2025.

\bibitem[Zhang et~al.(2020)Zhang, He, Sra, and Jadbabaie]{clipping}
Jingzhao Zhang, Tianxing He, Suvrit Sra, and Ali Jadbabaie.
\newblock Why gradient clipping accelerates training: {A} theoretical justification for adaptivity.
\newblock In \emph{8th International Conference on Learning Representations, {ICLR} 2020, Addis Ababa, Ethiopia, April 26-30, 2020}. OpenReview.net, 2020.
\newblock \url{https://openreview.net/forum?id=BJgnXpVYwS}.

\bibitem[Zhang et~al.(2023)Zhang, Li, Zong, Ying, He, and Qiu]{gaokao}
Xiaotian Zhang, Chunyang Li, Yi~Zong, Zhengyu Ying, Liang He, and Xipeng Qiu.
\newblock Evaluating the performance of large language models on {GAOKAO} benchmark.
\newblock \emph{CoRR}, abs/2305.12474, 2023.
\newblock \doi{10.48550/ARXIV.2305.12474}.
\newblock \url{https://doi.org/10.48550/arXiv.2305.12474}.

\bibitem[Zhong et~al.(2024)Zhong, Cui, Guo, Liang, Lu, Wang, Saied, Chen, and Duan]{agieval}
Wanjun Zhong, Ruixiang Cui, Yiduo Guo, Yaobo Liang, Shuai Lu, Yanlin Wang, Amin Saied, Weizhu Chen, and Nan Duan.
\newblock Agieval: {A} human-centric benchmark for evaluating foundation models.
\newblock In Kevin Duh, Helena G{\'{o}}mez{-}Adorno, and Steven Bethard, editors, \emph{Findings of the Association for Computational Linguistics: {NAACL} 2024, Mexico City, Mexico, June 16-21, 2024}, pages 2299--2314. Association for Computational Linguistics, 2024.
\newblock \doi{10.18653/V1/2024.FINDINGS-NAACL.149}.
\newblock \url{https://doi.org/10.18653/v1/2024.findings-naacl.149}.

\end{thebibliography}
